\documentclass[11pt]{article}
\usepackage[authoryear]{natbib}
\usepackage{amssymb}
\usepackage{amsfonts}
\usepackage{amsmath}
\usepackage{dsfont}
\usepackage{soul}
\usepackage{graphicx}
\usepackage{amsthm}
\usepackage{color}
\usepackage{comment}
\usepackage{setspace}
\usepackage{framed}
\usepackage{enumitem}
\usepackage{todonotes}
\usepackage{tikz}
\usepackage{hyperref}
\usepackage{rotating}
\usepackage{subfigure}
\usepackage[flushleft]{threeparttable}
\usepackage{multirow}
\newcommand{\cref}[2][1]{{\textup{(\hyperref[#2]{\ref*{#2}$_{#1}$})}}}
\usepackage{xr}
\usepackage{bm}
\usepackage[toc,page,header]{appendix}
\usepackage{minitoc}
\usepackage{booktabs}
\usepackage{geometry}
\title{Fast and Robust Online Inference \\ with Stochastic Gradient Descent via Random Scaling}

\author{
Sokbae  Lee\thanks{%
Address: 420 West 118th Street,  New York, NY 10027, USA. E-mail: \texttt{sl3841@columbia.edu}.} \\ \footnotesize Columbia University  \and
Yuan Liao\thanks{Address: 75 Hamilton St., New Brunswick, NJ 08901, USA. Email:
\texttt{yuan.liao@rutgers.edu}.}\\  \footnotesize   Rutgers University 
\and Myung Hwan Seo\thanks{%
Address: 1 Gwanak-ro, Gwanak-gu, Seoul 08826, Korea. E-mail: 
\texttt{myunghseo@snu.ac.kr}.} \\  \footnotesize  Seoul National University\and
Youngki  Shin\thanks{Address: 1280 Main St.\ W.,\ Hamiloton, ON L8S 4L8, Canada. Email:
\texttt{shiny11@mcmaster.ca}.}\\  \footnotesize   McMaster University
}

\usepackage[linesnumbered,ruled]{algorithm2e}
\SetKwInput{KwInput}{Input}                
\SetKwInput{KwOutput}{Output}              

\theoremstyle{plain}
\newtheorem{thm}{\protect\theoremname}
\newtheorem{asm}{Assumption}
\providecommand{\theoremname}{Theorem}

\begin{document}

\maketitle

\begin{abstract}
We develop a new method of online inference for a vector of parameters estimated by the Polyak-Ruppert averaging procedure of stochastic gradient descent (SGD) algorithms. We leverage insights from time series regression in econometrics and construct asymptotically pivotal statistics via random scaling. Our approach is fully operational with online data and is rigorously underpinned by a functional central limit theorem. Our proposed inference method has a couple of key advantages over the existing methods. First, the test statistic is computed in an online fashion with only SGD iterates and the critical values can be obtained without any resampling methods, thereby allowing for efficient implementation suitable for massive online data. Second, there is no need to estimate the asymptotic variance and our inference method is shown to be  robust  to changes in the tuning parameters for SGD algorithms in simulation experiments with synthetic data. \end{abstract}

\section{Introduction}\label{sec:intro}

We consider an inference problem for a vector of parameters defined by
\[
\beta^{*}:=\arg\min_{\beta\in\mathbb{R}^{d}}Q\left(\beta\right),
\]
where $Q\left(\beta\right):=\mathbb{E} \left[ q\left(\beta,Y\right) \right]$ is a real-valued population
objective function, $Y$ is a random vector,
and
$\beta \mapsto q\left(\beta,Y\right)$ is convex.
 For a given sample $\left\{ Y_{t}\right\} _{t=1}^{n}$,
let $\beta_{t}$ denote  
the stochastic gradient descent (SGD) solution path, that is, for each $t \geq 1$,  
 \begin{equation}
\beta_{t}=\beta_{t-1}-\gamma_{t}\nabla q\left(\beta_{t-1},Y_{t}\right),\label{eq:SGD1}
\end{equation}
 where $\beta_0$ is the initial starting value, 
 $\gamma_{t}$ is a step size, and
  $\nabla q\left(\beta_{t-1},Y_{t}\right)$ denotes the gradient
of $q\left(\beta,Y\right)$ with respect to $\beta$ at $\beta=\beta_{t-1}$. 
 We study the classical 
 \citet{Polyak1990}-\citet{ruppert1988efficient} averaging
 estimator 
$\bar{\beta}_{n}:=n^{-1}\sum_{t=1}^{n}\beta_{t}$.
\citet{polyak1992acceleration} 
established regularity conditions under which 
the averaging estimator $\bar{\beta}_{n}$ is asymptotically normal:
\begin{align*}
\sqrt{n}\left(\bar{\beta}_{n}-\beta^{*}\right)\overset{d}{\to}
\mathcal{N} (0, \Upsilon),
\end{align*}
where
the asymptotic variance $\Upsilon$ has a sandwich form $\Upsilon := H^{-1}S H^{-1}$, 
$H := \nabla^{2}Q\left(\beta^{*}\right)$ is the Hessian matrix
and $S := \mathbb{E} \left[\nabla q\left(\beta^{*},Y\right)\nabla q\left(\beta^{*},Y\right)'\right]$ is the score variance. 
The  Polyak-Ruppert estimator $\bar{\beta}_{n}$ can be computed recursively
by the updating rule $\bar{\beta}_{t}=\bar{\beta}_{t-1}\frac{t-1}{t}+\frac{\beta_{t}}{t}$,
which implies that it is well suited to the online setting.

Although the celebrated asymptotic normality result \citep{polyak1992acceleration}  was established about three decades ago, it is only past several years that online inference with $\bar{\beta}_{n}$ has gained increasing interest in the literature. 
It is challenging to estimate the asymptotic variance $\Upsilon$  in an online fashion.
This is because
the naive implementation of estimating it requires storing all data, thereby losing the advantage of online learning. In the seminal work of \citet{chen2020statistical}, the authors addressed this issue by  estimating   $H$ and $S$ using the online iterated estimator $\beta_t$, and recursively updating   them   whenever a new observation  is available.  They called this method a \textit{plug-in} estimator and showed that it consistently estimates the asymptotic variance and is ready for inference.  However, the plug-in estimator requires that the Hessian matrix be computed to estimate $H$.
In other words, it is necessary to have strictly more inputs than  the SGD solution paths $\beta_t$ to carry out inference. In applications, it can be demanding to compute the Hessian matrix. As an alternative, \citet{chen2020statistical}  proposed a \textit{batch-means} estimator that avoids separately estimating   $H^{-1}$ or $S$. This method directly estimates the variance of the averaged online estimator $\bar{\beta}_{n}$ 
by dividing $\{\beta_1,...,\beta_n\}$ into batches with increasing
batch size. The batch-means estimator is based on the idea that correlations among batches that are far apart decay exponentially fast; therefore, one can use nonparametric empirical covariance to estimate $\Upsilon$.
Along this line, \citet{zhu2020fully} extended the batch-means approach to allowing for  real-time recursive updates, which is desirable in the online setting.

The batch-means method produces batches of streaming samples, so that
data are weakly correlated when batches far apart. The distance between
batches is essential to control dependence among batches so it should be
chosen very carefully. In applications, we need to specify a sequence
which determines the batch size as well as the speed at which dependence
among batches diminish. While this is a new sequence one needs to tune,
it also affects the rate of convergence of the estimated covariance
matrix. As is shown by \citet{zhu2020fully}, the optimal choice of this sequence and
the batch size is related to the learning rate and could be very slow.
\citet{zhu2020fully} showed that the batch-mean covariance estimator converges no
faster than $O_P(n^{-1/4})$. Simulation results in both \citet{zhu2020fully} and this
paper show that indeed the coverage probability converges quite slowly.

Instead of estimating the asymptotic variance, \citet{fang2018online} proposed a  bootstrap procedure  for  online inference. 
Specifically, they proposed to use a large number (say, $B$) of randomly perturbed 
SGD solution paths: for all $b = 1, \ldots, B$, starting with $\beta_0^{(b)} = \beta_0$ and then iterating
\begin{equation}
\beta_{t}^{(b)}=\beta_{t-1}^{(b)}-\gamma_{t} \eta_t^{(b)} \nabla q\left(\beta_{t-1}^{(b)},Y_{t}\right),
\label{eq:SGD1:bt}
\end{equation}
where $\eta_t^{(b)} > 0$ is an independent and identically distributed random variable that has mean one and variance one.
The bootstrap procedure needs strictly more inputs than computing $\bar{\beta}_{n}$ 
and can be time-consuming.

In this paper, we propose a novel method of online inference for $\beta^{*}$. While the batch-means estimator aims to mitigate the effect of dependence among the averages of SGD iterates, on the contrary, we embrace dependence among them and propose to build a test statistic via random scaling.     We leverage insights from time series regression in econometrics
\citep[e.g.,][]{kiefer2000simple}
 and use a random transformation of $\beta_t$'s to construct asymptotically pivotal statistics. 
Our approach does  \textit{not} attempt to estimate the asymptotic variance $\Upsilon$,  but studentize $\sqrt{n}\left(\bar{\beta}_{n}-\beta^{*}\right)$ via
\begin{align}\label{def:random-scaling}
\widehat{V}_{n} := \frac{1}{n}\sum_{s=1}^{n} 
\left\{ \frac{1}{\sqrt{n}} \sum_{t=1}^{s} \left( \beta_{t}-\bar{\beta}_{n} \right) \right \} 
\left\{ \frac{1}{\sqrt{n}} \sum_{t=1}^{s} \left( \beta_{t}-\bar{\beta}_{n} \right) \right \}'.
\end{align}
The resulting statistic is not asymptotically normal but \emph{asymptotically pivotal} in the sense that its asymptotic distribution is free of any unknown nuisance parameters; thus, its critical values can be easily tabulated. 
Furthermore, the random scaling quantity $\widehat{V}_{n}$ does not require any additional inputs other than SGD paths $\beta_t$ and can be updated recursively. 
As a result, our proposed inference method has a couple of key advantages over the existing methods. First, the test statistic is computed in an online fashion with only SGD iterates and the critical values can be obtained without any resampling methods, thereby allowing for efficient implementation suitable for massive online data. Second, there is no need to estimate the asymptotic variance and our inference method is shown to be  robust  to changes in the tuning parameters for SGD algorithms in simulation experiments with synthetic data.   

\paragraph{Related work on SGD}
The SGD methods, which are  popular in the setting of online learning
\citep[e.g.,][]{hoffman2010online,mairal2010online},  
have been studied extensively in the recent decade. Among other things, probability bounds on  statistical errors  have been derived. For instance,
\citet{bach2013non} showed that for both the square loss and   the logistic loss, one  may use the smoothness of the
loss function to obtain algorithms that have a fast convergence rate without any strong convexity. 
See \citet{rakhlin2011making} and \citet{hazan2014beyond} for related results on convergence rates.
\citet{duchi2011adaptive} proposed AdaGrad, employing the  square root of the inverse diagonal Hessian matrix to adaptively control the gradient steps of SGD and derived regret bounds for the  loss function.   \citet{kingma2014adam} introduced Adam, computing  adaptive learning rates for different parameters from estimates of first and second moments of the gradients.   \citet{liang2017statistical} employed a similar idea as AdaGrad to adjust the gradient direction and showed that the  distribution for inference can be simulated iteratively.
\citet{Toulis2017} developed implicit SGD procedures and established the resulting estimator's asymptotic normality.
\citet{pmlr-v99-anastasiou19a}
and
\citet{pmlr-v125-mou20a}
developed 
some  results for non-asymptotic inference. 
 
 
 \paragraph{Related work in econometrics}
The random scaling by means of the partial sum process of the same summands has been actively employed to estimate the so-called long-run variance, which is the sum of all the autocovariances, in the time series econometrics since it was suggested by \citet{kiefer2000simple}. 
The literature has documented evidences that the random scaling stabilizes the excessive finite sample variation in the traditional consistent estimators of the long-run variance; see, e.g., \citet{velasco2001edgeworth}, \citet{sun2008optimal}, and a recent review in \citet{lazarus2018har}. 
The insight has proved valid in broader contexts, where the estimation of the asymptotic variance is challenging, 
such as in \citet{kim2011spatial}, which involves spatially dependent data,  and
\citet{gupta2021robust} for a high-dimensional inference. 
We show in this paper that it is indeed useful in on-line inference, which has not been explored to the best of our knowledge.
{\color{black} While our experiments focus on one of the earlier proposals of the random scaling methods, there are numerous alternatives, see e.g. \citet{sun2014fixed}, which warrants future research on the optimal random scaling method.}

\paragraph{Notation}

Let $a'$ and $A'$, respectively, denote the transpose of vector $a$ and matrix $A$. 
Let $| a |$ denote the Euclidean norm of vector $a$
and $\| A \|$ the Frobenius norm of matrix $A$. 
Also, let $\ell^{\infty}\left[0,1\right]$ denote the  set of bounded continuous functions on $[0,1].$






\section{Online Inference}\label{sec:theory}

In this section, we 
first present asymptotic theory that underpins our inference method
and describe our proposed online inference algorithm.
Then, we explain our method in comparison with the existing methods using the linear regression model as an example.  

\subsection{Functional central limit theorem for online SGD}

We first  extend
\citet{polyak1992acceleration}'s central limit theorem (CLT) to a \emph{functional} CLT (FCLT), that is,
\begin{equation}\label{eq4}
\frac{1}{\sqrt{n}}\sum_{t=1}^{\left[nr\right]}\left(\beta_{t}-\beta^{*}\right)\Rightarrow \Upsilon^{1/2}W\left(r\right),\quad r\in\left[0,1\right],
\end{equation}
where $\Rightarrow$ stands for the weak convergence in $\ell^{\infty}\left[0,1\right]$
and $W\left(r\right)$ stands for a vector of the independent standard
Wiener processes on $\left[0,1\right]$. That is, the partial sum of the online updated estimates $\beta_t$ converges weakly to a rescaled Wiener process, with scaling equal to the square root  asymptotic variance of the Polyak-Ruppert average. 
The CLT proved in \citet{polyak1992acceleration}  is then
a special case with $r=1$. Building on this extension, we propose
an online inference procedure. Specifically, 
using the random scaling matrix $\widehat{V}_n$ defined in \eqref{def:random-scaling}, we consider 
the following t-statistic
\begin{align}\label{t-stat}
\frac{\sqrt{n}\left(\bar{\beta}_{n,j}-\beta_{j}^{*}\right)}{\sqrt{\widehat{V}_{n,jj}}},
\end{align}
where the subscripts $j$ and $jj$, respectively, denote the $j$-th element of a vector 
and the $(j,j)$ element of a matrix.  
Then, the FCLT yields that the t-statistic 
is asymptotically pivotal. Note that instead of using an estimate of $\Upsilon$, we use the random scaling $\widehat V_{n}$ for the proposed t-statistic. As a result, the limit is not conventional standard
normal but a mixed normal. It can be utilized to construct confidence
intervals for $\beta_{j}^{*}$ for each $j$. A substantial advantage
of this random scaling is that it does not have to estimate an
analytic asymptotic variance formula and tends to be more robust in finite
samples. As mentioned in the introduction, this random scaling idea has been widely used in the literature
known as the fixed bandwidth heteroskedasticity and autocorrelation robust (HAR) inference \citep[e.g.,][]{kiefer2000simple,lazarus2018har}. 

More generally, for any $\ell \leq d$ linear restrictions
\[
H_{0}: R\beta^{*} = c,
\]
where $R$ is  an $(\ell \times d)$-dimensional known matrix of  rank $\ell$
and
$c$ is an $\ell$-dimensional known vector,
the conventional Wald test based
on $\widehat{V}_{n}$ becomes asymptotically pivotal. 
To establish this result formally, we make the following assumptions \`{a} la \citet{polyak1992acceleration}.

\begin{asm}
    \label{ass:PJ's Assumption 3}
\begin{itemize} 
    \item[(i)] There exists a function $\Psi(\beta):\mathbb{R}^{d}\rightarrow\mathbb{R}$
    such that for some $\lambda>0,\:\alpha>0,\:\varepsilon>0,\:L>0$,
    and all $x,y\in\mathbb{R}^{d}$, $\Psi(x)\geq\alpha|x|^{2}$, $|\nabla \Psi(x)-\nabla \Psi(y)|\leq L|x-y|$,
    $\Psi(\beta^{*})=0$, and $\nabla \Psi(\beta-\beta^{*})^{T}\nabla Q(\beta)>0$
    for $\beta\neq\beta^{*}$ hold true. Moreover, $\nabla \Psi(\beta-\beta^{*})^{T}\nabla Q(\beta)\geq\lambda \Psi(\beta)$
    for all $|\beta-\beta^{*}|\leq\varepsilon$.
    \item[(ii)] The Hessian matrix $H$ is positive definite and there exist $K_{1}<\infty$, $\varepsilon>0$, $0<\lambda\leq1$
    such that $|\nabla Q(\beta)-H(\beta-\beta^{*})|\leq K_{1}|\beta-\beta^{*}|^{1+\lambda}$,
    for all $|\beta-\beta^{*}|\leq\varepsilon$. 
    
    \item[(iii)] The sequence $\left\{ \xi_{t}:=\nabla Q\left(\beta_{t-1}\right)-\nabla q\left(\beta_{t-1},Y_{t}\right) \right\}_{t\geq1}$
    is a martingale-difference sequence (mds), defined on a probability space
    $(\Omega,\mathcal{{F}},\mathcal{{F}}_{t},P)$, i.e., $E(\xi_{t}|\mathcal{{F}}_{t-1})=0$
    almost surely, and for some $K_{2}$, $E(|\xi_{t}|^{2}|\mathcal{{F}}_{t-1})+|\nabla Q(\beta_{t-1})|^{2}\leq K_{2}(1+|\beta_{t-1}|^{2})$
    a.s. for all $t\geq1$. Then, the following decomposition takes place:
    $\xi_{t}=\xi_{t}(0)+\zeta_{t}(\beta_{t-1})$, where $\mathbb{E} (\xi_{t}(0)|\mathcal{{F}}_{t-1})=0$
    a.s., $\mathbb{E} (\xi_{t}(0)\xi_{t}(0)'|\mathcal{{F}}_{t-1})\xrightarrow{P}S$
    as $t\rightarrow\infty$; $S>0$ ($S$ is symmetrical and positive
    definite), $\sup\limits _{t}\mathbb{E} (|\xi_{t}(0)|^{2}I(|\xi_{t}(0)|>C|\mathcal{\mathcal{{F}}}_{t-1})\xrightarrow{P}0$
    as $C\rightarrow\infty$, and for all $t$ large enough, $\mathbb{E} (|\zeta_{t}(\beta_{t-1})|^{2}|\mathcal{{F}}_{t-1})\leq\delta(\beta_{t-1})$
    a.s. with $\delta(\beta)\rightarrow0$ as $\beta\rightarrow0$.
    
    \item[(iv)] It holds that $\gamma_{t}=\gamma_0 t^{-a} $ for some $1/2 < a <1$.
\end{itemize}   
\end{asm}

\begin{asm}
\label{as2}
\begin{itemize}
\item[(i)]
Let $\mathcal{F}_{s}^{t}$ denote the sigma field
generated by $\left\{ \xi_{s},...,\xi_{t}\right\} $. If $X\in\mathcal{F}_{0}^{t}$
and $Y\in\mathcal{F}_{t+d}^{\infty}$, then $\left\Vert \mathbb{E}X'Y-\mathbb{E}X'\mathbb{E}Y\right\Vert \leq c\left(\left|d\right|^{-\eta}\right)$
for some $\eta>1$ and $c>0$. 
\item[(ii)] 
For $p\geq2/\left(1-a\right)$, $\mathbb E\left\Vert \xi_{t}\right\Vert ^{p}$
is bounded. 
\end{itemize}

\end{asm}

Assumptions~\ref{ass:PJ's Assumption 3} (i)--(iii) are identical to Assumptions 3.1--3.3  of \citet{polyak1992acceleration} and Assumption 1 (iv) is the standard learning rate. 
Assumption \ref{as2} adds mixing-type condition and the bounded moments condition to enhance the results for uniform convergence, which is needed to prove the functional CLT.

Given these assumptions, The following theorem is a formal statement under the conditions stated above. Note that the proof of the FCLT requires bounding some processes, indexed by $r$,  uniformly over $r$. Our proof is new, as some of these processes \textit{cannot} be written as  partial sums of martingale  differences. Hence we cannot simply apply results such as Doob's inequalities.  Recent works, as in Zhu and Dong (2019), developed FCLT using bounded sequences. Our proof extends theirs to possibly unbounded sequences but with finite moments, and uses  new technical arguments.

\begin{thm}
\label{thm:Wald} Suppose rank$(R)=\ell$. Under Assumption \ref{ass:PJ's Assumption 3} and $H_{0}$,
\begin{eqnarray*}
&&n\left(R\bar{\beta}_{n}-c\right)'\left(R\widehat{V}_{n}R'\right)^{-1}\left(R\bar{\beta}_{n}-c\right)\cr
&&\overset{d}{\to}W\left(1\right)'\left(\int_{0}^{1}\bar{W}(r)\bar{W}(r)'dr\right)^{-1}W\left(1\right),
\end{eqnarray*}
where $W$ is an $\ell$-dimensional vector of the standard Wiener
processes and $\bar{W}\left(r\right):=W\left(r\right)-rW\left(1\right)$. 
\end{thm}

\begin{proof}[Proof of Theorem \ref{thm:Wald}] 
Rewrite (\ref{eq:SGD1})
as 
\begin{equation}
\beta_{t}=\beta_{t-1}-\gamma_{t}\nabla Q\left(\beta_{t-1}\right)+\gamma_{t}\xi_{t}\label{eq:SGD2}.
\end{equation}
Let $\Delta_{t}:=\beta_{t}-\beta^{*}$
and $\bar{\Delta}_{t}:=\bar{\beta}_{t}-\beta^{*}$ to denote the errors
in the $t$-th iterate and that in the average estimate at $t$, respectively.
Then, subtracting $\beta^{*}$ from both sides of (\ref{eq:SGD2})
yields that
\[
\Delta_{t}=\Delta_{t-1}-\gamma_{t}\nabla Q\left(\beta_{t-1}\right)+\gamma_{t}\xi_{t}.
\]
Furthermore, for $r\in\left[0,1\right]$, introduce a partial sum process
\[
\bar{\Delta}_{t}\left(r\right):=t^{-1}\sum_{i=1}^{\left[tr\right]}\Delta_{i},
\]
whose weak convergence we shall establish. 

Specifically, 
we extend Theorem 2 in \citet[PJ hereafter]{polyak1992acceleration} to an FCLT. The first step is a uniform approximation of the partial sum process to another partial sum process $\bar{\Delta}_{t}^{1}\left(r\right)$
of 
\[
\Delta_{t}^{1}:=\Delta_{t-1}^{1}-\gamma_{t}H\Delta_{t-1}^{1}+\gamma_{t}\xi_{t}\quad\text{and}\quad\Delta_{0}^{1}=\Delta_{0}.
\]
That is, we need to show that $\sqrt{t}\sup_{r}\left|\bar{\Delta}_{t}\left(r\right)-\bar{\Delta}_{t}^{1}\left(r\right)\right|=o_{p}\left(1\right)$.
According to Part 4 in the proof of PJ's Theorem 2, this is indeed
the case. 

Turning to the weak convergence of $\sqrt{t}\bar{\Delta}_{t}^{1}\left(r\right)$,
we extend PJ's Theorem 1. Following its decomposition in (A10), write
\[
\sqrt{t}\bar{\Delta}_{t}^{1}\left(r\right)=I^{\left(1\right)}\left(r\right)+I^{\left(2\right)}\left(r\right)+I^{\left(3\right)}\left(r\right),
\]
where 
\begin{align*}
I^{\left(1\right)}\left(r\right) & :=\frac{1}{\gamma_{0}\sqrt{t}}\alpha_{\left[tr\right]}\Delta_{0},\\
I^{\left(2\right)}\left(r\right) & :=\frac{1}{\sqrt{t}}\sum_{j=1}^{\left[tr\right]}H^{-1}\xi_{j},\\
I^{\left(3\right)}\left(r\right) & :=\frac{1}{\sqrt{t}}\sum_{j=1}^{\left[tr\right]}w_{j}^{\left[tr\right]}\xi_{j},
\end{align*}
where $\alpha_{t}=\left(t\gamma_{t}\right)^{-1}\leq K$ and $\left\{ w_{j}^{\left[tr\right]}\right\} $
is a bounded sequence such that $t^{-1}\sum_{j=1}^{t}\left\Vert w_{j}^{t}\right\Vert \to0$.
Then, $\sup_{r}\left\Vert I^{\left(1\right)}\left(r\right)\right\Vert =o_{p}\left(1\right)$. 
Suppose for now that $\mathbb E\sup_r \| I^{(3)}\|^p=o(1)$ for some $p\geq 1$. The bound for $I^{(3)}$ requires sophisticated arguments, as $w_j\xi_j$ is \textit{not} mds, even though $\xi_j$ is. So we  develop new technical arguments to bound this term, whose proof is left at the end of the proof.

Then the FCLT for mds, see
e.g. Theorem 4.2 in \citet{Hall-Heyde},  applies to $I^{\left(2\right)}\left(r\right)$,
whose regularity conditions are verified in the proof (specifically, Part 1) of
the theorem in PJ  to apply the mds CLT for $I^{\left(2\right)}\left(1\right)$. 
This shows that $I^{(2)}$ converges weakly to a rescaled Wiener process $\Upsilon^{1/2}W\left(r\right)$. This establishes the FCLT in (\ref{eq4}).  

Now let $C_n(r):=R\frac{1}{\sqrt{n}}\sum_{t=1}^{\left[nr\right]}\left(\beta_{t}-\beta^{*}\right)$. Also let $\Lambda= (R\Upsilon R')^{1/2}$, which exists and is invertible as long as $l\leq d$. 
(\ref{eq4}) then shows that  for some vector of independent standard Wiener process $W^*(r)$,
$$
C_n(r)\Rightarrow \Lambda W^*(r).
$$
In addition,
$R\widehat V_nR'= \frac{1}{n}\sum_{s=1}^n[C_n(\frac{s}{n})-\frac{s}{n}C_n(1)][C_n(\frac{s}{n})-\frac{s}{n}C_n(1)]'$, 
where the sum is also an integral over $ r $ as $ C_n(r) $ is a partial sum process, 
and  $R(\bar\beta_n-\beta^*)=\frac{1}{\sqrt{n}}C_n(1)$. Hence
 $
n\left(R\bar{\beta}_{n}-c\right)'\left(R\widehat{V}_{n}R'\right)^{-1}\left(R\bar{\beta}_{n}-c\right) $ is a continuous functional of $C_n(\cdot)$.  The desired result in the theorem then follows
from the continuous mapping theorem. 

It remains to bound $I^{(3)}(r)$ uniformly. Let $S_{t}=\sum_{j=1}^{t-1}w_{j}^{t}\xi_{j}$.
Then, 
\[
\mathbb{E}\sup_{r}\left\Vert I^{\left(3\right)}\left(r\right) \right\Vert ^{p}\leq t^{-p/2}
\mathbb{E}\sup_{r}\left\Vert S_{\left[tr\right]}\right\Vert ^{p}
\leq  t^{-p/2}\sum_{m=1}^{t}\mathbb{E}\left\Vert S_{m}\right\Vert ^{p}
\]
and 
\[
\mathbb{E}\left\Vert S_{m}\right\Vert ^{p}=\sum_{j_{1},...,j_{p}=1}^{m-1}w_{j_{1}}^{m}\cdots w_{j_{p}}^{m}\mathbb{E}\xi_{j_{1}}'\cdots\xi_{j_{p}}.
\]
Note that $\mathbb{E}\xi_{j_{1}}'\cdots\xi_{j_{p}}=0$ for distinct
values of $j_{1},...,j_{p}$ as $\xi_{j}$ is mds. When $\left(j_{1},...,j_{p}\right)\in A_k$,  where set $A_k$ means the collection of indices that consist of vectors of $k$ distinct values from $1,...,m-1$,
\[
\sum_{\left(j_{1},...,j_{p}\right)\in A_k}w_{j_{1}}^{m}\cdots w_{j_{p}}^{m}\mathbb{E}\xi_{j_{1}}'\cdots\xi_{j_{p}}=O\left(\left(\sum_{j=1}^{m-1}\left\Vert w_{j}^{m}\right\Vert \right)^{k}\right)
\]
since $\mathbb{E}\xi_{j_{1}}'\cdots\xi_{j_{p}}$ is bounded and $\sum_{j=1}^{m}\left\Vert w_{j}^{m}\right\Vert ^{b}=O\left(\sum_{j=1}^{m}\left\Vert w_{j}^{m}\right\Vert \right)$
for any $b$ due to the boundedness of $\left\Vert w_{j}^{m}\right\Vert $. 
According to Lemma 2 in \citet{Zhu:Dong}, $\sum_{j=1}^{m}\left\Vert w_{j}^{m}\right\Vert =o\left(m^{a}\right)$.
Hence
\begin{equation}\label{eq:boundloos}
\sum_{\left(j_{1},...,j_{p}\right)\in A_k}w_{j_{1}}^{m}\cdots w_{j_{p}}^{m}\mathbb{E}\xi_{j_{1}}'\cdots\xi_{j_{p}}\leq o(m^{ak})
\end{equation}
which holds uniformly for $m, k$. For notational simplicity, write
$$
K(m):=w_{j_{1}}^{m}\cdots w_{j_{p}}^{m}\mathbb{E}\xi_{j_{1}}'\cdots\xi_{j_{p}}
$$
Thus
\begin{align*}
&   \mathbb{E}\sup_{r}\left\Vert I^{\left(3\right)}\left(r\right) \right\Vert ^{p}\cr
&\leq t^{-p/2}\sum_{m=1}^t\sum_{k\leq \min\{m-1,p-1\}}\sum_{j_{1},...,j_{p}\in A_k,\leq m-1 } K(m)\cr
&\leq B_1+ B_2\cr
B_1&:= t^{-p/2}\sum_{m=1}^t\sum_{k\leq \min\{m-1,p/2\}}\sum_{j_{1},...,j_{p}\in A_k,\leq m-1 } K(m)\cr
&\leq o(1)
t^{-p/2}\sum_{m=1}^t\sum_{k\leq \min\{m-1,p/2\}} m^{ak}
\cr
&\leq o(1)   t^{1+ap/2-p/2} =o(1)\cr
B_2&:= t^{-p/2}\sum_{m=1}^t\sum_{p/2<k\leq \min\{m-1,p-1\}}\sum_{j_{1},...,j_{p}\in A_k,\leq m-1 } K(m),
\end{align*}
where for $B_1$, $1+ap/2-p/2 <0$ because $1< (1-a)p/2$.

 We now examine $B_2$, which is the case  $k>p/2$ in the subindex.  We can tighten the bound (\ref{eq:boundloos}) on $\mathbb{E}\xi_{j_{1}}'\cdots\xi_{j_{p}}$ 
using the mixing condition in Assumption \ref{as2}. The proof holds for a general $p$, but without loss of generality, we take $p=6$ as an example. Then $k\in\{4,5\}$. 
 Now suppose $k=4$. There
are two types. One type is that $j_{1}=j_{2}$, $j_{3}=j_{4}$, and
$j_{5}\neq j_{6}$ and $j_{5}$ and $j_{6}$ are also different from
$j_{1}$and $j_{3}$. The other type is where $j_{1}=j_{2}=j_{3}$
and the other elements are distinct. We only consider $j_{1}=j_{2}$>
$j_{3}=j_{4}$> $j_{5}>j_{6}$ explicitly as the other cases are similar.
Note that 
\begin{eqnarray*}
\left|\mathbb{E}\xi_{j_{1}}'\cdots\xi_{j_{p}}\right|&\leq&\left|\mathbb{E}\xi_{j_{1}}'\xi_{j_{1}}\xi_{j_{3}}'\xi_{j_{3}}E\xi_{j_{5}}'\xi_{j_{6}}\right|+O\left(|j_{3}-j_{5}|^{-\eta}\right)\cr
&=&O\left(|j_{3}-j_{5}|^{-\eta}\right),
\end{eqnarray*}
 due to the mixing-condition and the mds property of $\xi_{t}$.
And as $\left\Vert w_{j}^{t}\right\Vert $ is bounded and the sequence
$j^{-\eta}$ is summable for $\eta>1$, we conclude, uniformly in $m$,
\begin{align*}
\sum_{j_{1}=j_{2}>j_{3}=j_{4}>j_{5}>j_{6}} K(m)&=O\left(\left(\sum_{j=1}^{m-1}\left\Vert w_{j}^{m}\right\Vert \right)^{p/2}\right) \\
&=o(m^{ap/2}).
\end{align*}
Similarly we reach the same bound for $k=5$. The case with larger
$p$ also proceeds similarly to conclude that 
$$
B_2\leq   o(t^{ap/2+1-p/2}) =o(1).
$$
This concludes that $\mathbb{E}\sup_{r}\left\Vert I^{\left(3\right)}\left(r\right) \right\Vert ^{p}= o(1)$.
 
\end{proof}

As an important special case of Theorem \ref{thm:Wald}, the t-statistic defined in \eqref{t-stat} converges in distribution to the following pivotal limiting distribution:
for each $j = 1,\ldots,d$, 
\begin{align}\label{t-stat-limit}
\frac{\sqrt{n}\left(\bar{\beta}_{n,j}-\beta_{j}^{*}\right)}{\sqrt{\widehat{V}_{n,jj}}}
\overset{d}{\to}
W_1\left(1\right)  \left[ \int_{0}^{1} \left\{ W_1\left(r\right)-rW_1\left(1\right) \right\}^2 dr\right]^{-1/2}, 
\end{align}
where 
 $W_1$ is a one-dimensional standard Wiener process.

\paragraph{Related work on \citet{polyak1992acceleration}}

There exist papers that have extended \citet{polyak1992acceleration}
to more general forms
\citep[e.g.,][]{Kushner:Yang:93,Godichon-Baggioni,Su:Zhu,Zhu:Dong}.
The stochastic process defined in \citet[equation (2.2)]{Kushner:Yang:93} is different from the partial sum process in \eqref{eq4}.
\citet{Godichon-Baggioni}
considers parameters taking values in a separable Hilbert space and as
such it considers a generalization of Polyak-Juditsky to more of an
empirical process type while our FCLT concerns the partial sum
processes.
\citet{Su:Zhu}'s HiGrad tree divide updates into levels, with the idea that
correlations among distant SGD iterates decay rapidly. Their Lemma 2.5
considers the joint asymptotic normality of certain $K$ partial sums for
a finite $K$ while our FCLT is for the partial sum process indexed by
real numbers on the $\left[0,1\right]$ interval. 
\citet{Zhu:Dong} appears closer than the others to
our FCLT for the partial sums, although the set of sufficient conditions is not the same as ours. However, we emphasize that the more innovative
part of our work is the way how we utilize the FCLT than the FCLT
itself.
Indeed, it appears that prior to this paper, there is no other work in the literature that makes use of the FCLT as this paper does in order to conduct on-line inference with SGD.

\subsection{An Algorithm for Online Inference}

Just as the average SGD estimator can be updated recursively via  $\bar{\beta}_{t}=\bar{\beta}_{t-1}\frac{t-1}{t}+\frac{\beta_{t}}{t}$,
the statistic $\widehat{V}_{t}$ can also be updated in an online fashion.
To state the online updating rule for $\widehat{V}_{t}$,
note that 
\begin{align*}
t^{2}\widehat{V}_{t}&=\sum_{s=1}^{t}\left(\sum_{j=1}^{s}\beta_{j}-s\bar{\beta}_{t}\right)\left(\sum_{j=1}^{s}\beta_{j}-s\bar{\beta}_{t}\right)' \\
&=\sum_{s=1}^{t}\sum_{j=1}^{s}\beta_{j}\sum_{j=1}^{s}\beta_{j}'
- \bar{\beta}_{t}\sum_{s=1}^{t}s\sum_{j=1}^{s}\beta_{j}' \\
&\;\;\; - \sum_{s=1}^{t}s\sum_{j=1}^{s}\beta_{j} \bar{\beta}_{t}' 
+\bar{\beta}_{t}\bar{\beta}_{t}'\sum_{s=1}^{t}s^{2}
\end{align*}
and 
\begin{align*}
\sum_{s=1}^{t}\sum_{j=1}^{s}\beta_{j}\sum_{j=1}^{s}\beta_{j}' & =\sum_{s=1}^{t-1}\sum_{j=1}^{s}\beta_{j}\sum_{j=1}^{s}\beta_{j}'+t^{2}\bar{\beta}_{t}\bar{\beta}_{t}'\\
\sum_{s=1}^{t}s\sum_{j=1}^{s}\beta_{j} & =\sum_{s=1}^{t-1}s\sum_{j=1}^{s}\beta_{j}+t^{2}\bar{\beta}_{t}.
\end{align*}
Thus, at step $t-1$, we only need to keep the three quantities, $\bar{\beta}_{t-1}$,
\[
A_{t-1}=\sum_{s=1}^{t-1}\sum_{j=1}^{s}\beta_{j}\sum_{j=1}^{s}\beta_{j}',\quad\text{and}\quad b_{t-1}=\sum_{s=1}^{t-1}s\sum_{j=1}^{s}\beta_{j},
\]
to update $\widehat{V}_{t-1}$ to $\widehat{V}_{t}$ using the new observation
$\beta_{t}$. 
The following algorithm summarizes the arguments above. 

\begin{algorithm}[htbp]
 \SetKwInput{KwInit}{Initialize}
 \SetKwInput{KwReceive}{Receive}

 \KwInput{function $q(\cdot)$, parameters $(\gamma_0, a)$ for step size $\gamma_t=\gamma_0 t^{-a}$ for $t \geq 1$}

 \KwInit{
  set initial values for $\beta_0, \bar{\beta}_0, A_0$
 }
 \For{$t = 1,2, \ldots $} {
    \KwReceive{new observation $Y_t$}
    $\beta_{t}=\beta_{t-1}-\gamma_{t}\nabla q\left(\beta_{t-1},Y_{t}\right)$

    $\bar{\beta}_{t}=\bar{\beta}_{t-1}\frac{t-1}{t}+\frac{\beta_{t}}{t}$   

    $A_{t} = A_{t-1} + t^{2}\bar{\beta}_{t}\bar{\beta}_{t}'$

    $b_{t} = b_{t-1} + t^{2}\bar{\beta}_{t}$

    Obtain $\widehat{V}_{t}$ by 
      \[
      \widehat{V}_{t}
      = t^{-2} \left( A_t - \bar{\beta}_{t} b_{t}' - b_{t}\bar{\beta}_{t}' +\bar{\beta}_{t}\bar{\beta}_{t}'\sum_{s=1}^{t}s^{2} \right)
      \]

    \KwOutput{$\bar{\beta}_t$, $\widehat{V}_{t}$}
 }

\caption{Online Inference with SGD via Random Scaling}\label{alg:sgd}
\end{algorithm}


Once $\bar{\beta}_{n}$ and $\widehat{V}_{n}$ are obtained, it is straightforward to carry out inference. For example,
we can use the t-statistic in \eqref{t-stat} to construct 
 the $(1-\alpha)$ asymptotic confidence interval for the $j$-th element $\beta_{j}^{*}$ of $\beta^{*}$ by 
\[
\left[
\bar{\beta}_{n,j} - \textrm{cv} (1- \alpha/2) \sqrt{\frac{\widehat{V}_{n,jj}}{n}}, \;
\bar{\beta}_{n,j} + \textrm{cv} (1- \alpha/2) \sqrt{\frac{\widehat{V}_{n,jj}}{n}} \;
\right],
\]
where the critical value $\textrm{cv} (1- \alpha/2)$ is tabulated in \citet[Table I]{Abadir:Paruolo:97}.
The limiting distribution in \eqref{t-stat-limit}  is mixed normal and symmetric around zero. 
For easy reference, we reproduce the critical values in Table~\ref{tab:cv}.
When $\alpha = 0.05$, the critical value is 6.747.
Critical values for testing linear restrictions $H_{0}: R\beta^{*} = c$
are given in \citet[Table II]{kiefer2000simple}.

\begin{table}[htbp]
\centering
\begin{threeparttable}
\caption{Asymptotic critical values of the t-statistic}\label{tab:cv}
\begin{tabular}{lcccc}
\hline
Probability  & 90\% & 95\% & 97.5\% & 99\% \\
Critical Value & 3.875 & 5.323 & 6.747 & 8.613 \\
\hline
\end{tabular}
 \begin{tablenotes}\small
\item Note. The table gives one-sided asymptotic critical values that satisfy 
$\mathrm{Pr}( \hat{t} \leq c ) = p$ asymptotically, where
$p \in \{0.9, 0.95, 0.975, 0.99\}$.
Source: \citet[Table I]{Abadir:Paruolo:97}. 
\end{tablenotes}
\end{threeparttable}
\end{table}%

\subsection{Estimation of the linear regression model}

In this subsection, we consider the least squares estimation of the linear regression model $y_{t}=x_{t}'\beta^*+\varepsilon_{t}$.
 In this example, the stochastic gradient sequence $\beta_t$ is given by 
 \[
 \beta_{t}=\beta_{t-1}-\gamma_{t}x_{t}\left(x_{t}'\beta_{t-1}-y_{t}\right),
 \]
 where $\gamma_{t}=\gamma_0 t^{-\alpha}$ with $1/2<\alpha<1$ is 
 the step size. 
 This linear regression model satisfies Assumption~\ref{ass:PJ's Assumption 3}, as shown by \citet[section 5]{polyak1992acceleration}.
The   asymptotic variance of $\bar\beta_n$ is $\Upsilon=H^{-1}SH^{-1}$ where $H= \mathbb E x_tx_t'$ and $S= \mathbb E x_tx_t'\epsilon_t^2$.   

Our proposed method would standardize using $\widehat V_n$ according to (\ref{def:random-scaling}), 
which does \textit{not} consistently estimate  $\Upsilon$.
We use  critical values as tabulated in Table~\ref{tab:cv}, whereas
the existing methods (except for the bootstrap) would seek for consistent estimation of  $\Upsilon.$ 
For instance, the plug-in method respectively estimates $H$ and $S$   by $\widehat H=\frac{1}{n}\sum_{t=1}^nx_tx_t'$, and 
$\widehat S=\frac{1}{n}\sum_{t=1}^n x_tx_t'\widehat\epsilon_t^2$, where $\widehat\epsilon_t=y_t-x_t'\beta_{t-1}.$ 
Note that    
$\widehat H^{-1}$ does not rely on the updated $\beta_t$ but may not be easy to compute if $\dim(x_t)$ is moderately large. 
Alternatively, the batch-mean method first splits the iterates  $\beta_t$'s into $M+1$ batches, 
discarding the first batch as the burn-in stage, and estimates $\Upsilon$ directly by 
\[
 \widehat\Upsilon_1=\frac{1}{M}\sum_{k=1}^Mn_k(\widehat\beta_k-\bar\beta_n)(\widehat\beta_k-\bar\beta_n)',
\]
 where $\widehat\beta_k$ is the mean of $\beta_t$'s for the $k$-th batch and $n_k$ is the batch size.
 One may also discard the first batch when calculating  $\bar\beta_n$ in $\widehat\Upsilon$. 
 As noted by \citet{zhu2020fully}, a serious drawback of this approach is  that one needs to know the total number
 of  $n$ as a priori, so one needs to recalculate $\widehat\Upsilon_1$ whenever a new observation arrives. Instead, \citet{zhu2020fully} proposed a ``fully online-fashion'' covariance estimator, which splits the iterates $\beta_t$ into $n$ batches $B_1,...,B_n$, and estimates the covariance by 
\[
 \widehat\Upsilon_2=\frac{1}{\sum_{k=1}^n |B_k|} \sum_{k=1}^n(S_k-|B_k| \bar \beta_n)(S_k-|B_k| \bar \beta_n)',
\]
 where $S_k$ denotes the sum of all elements in $B_k$ and $|B_k|$ denotes the size of the $k$-th batch. The batches are overlapped. For instance, fix a pre-determined sequence $\{1,3,5,7,...\}$, we can set 
 \begin{eqnarray*}
    B_1&=& \{\beta_1\}, \quad B_2=\{\beta_1, \beta_2\},  \cr
    B_3&=&\{\beta_3 \}, \quad B_4=\{\beta_3, \beta_4 \}, \cr
    B_5&=&\{\beta_5 \},\quad B_6=\{\beta_5, \beta_6 \},
\end{eqnarray*}
and subsequent $B_ t$'s are defined analogously. 
Our proposed scaling $\widehat V_n$ is similar to $\widehat\Upsilon_2$ in the sense that it    can be formulated as:
$$
\widehat V_n=\frac{1}{n^2}\sum_{k=1}^n(S_k-|B_k^*|\bar\beta_n) (S_k-|B_k^*|\bar\beta_n)'
$$
with particular choice of batches being:
\begin{eqnarray*}
B_1^*&=&\{\beta_1\}, B_2^*=\{\beta_1,\beta_2\}, B_3^*=\{\beta_1,\beta_2,\beta_3\},...,\cr
B_k^*&=&B_{k-1}^*\cup\{\beta_k\},...
\end{eqnarray*}
However, there is a key difference between $\widehat V_n$ and $\widehat\Upsilon_2$: the batches used by $\widehat\Upsilon_2$, though they can be overlapped,   are required to be weakly correlated as they become far apart. In contrast, $B_k^*$ are strongly correlated and strictly nested.  Thus, we embrace dependences among $B_k^*$, and reach the scaling $\widehat V_n$ that does not consistently estimate $\Upsilon.$
The important advantage of our approach is that there is no need to choose the batch size. 

In the next section, we provide   results of experiments that compare different methods in the linear regression model.

\section{Experiments}\label{sec:MC}


In this section we investigate the numerical performance of the random scaling method via Monte Carlo experiments. 
We consider two baseline models: linear regression and logistic regression. 
We use the Compute Canada Graham cluster composed of Intel CPUs (Broadwell, Skylake, and Cascade Lake at 2.1GHz--2.5GHz) and they are assigned with 3GB memory.

\paragraph{Linear Regression} The data are generated from 
\begin{align*}
y_{t} = x_t'\beta^* + \varepsilon_t~~\mbox{for}~~t=1,\ldots,n,
\end{align*}
where $x_t$ is a $d$-dimensional covariates generated from the multivariate normal distribution $\mathcal{N} (0,I_d)$, $\varepsilon_t$ is from $N(0,1)$, and $\beta^*$ is equi-spaced on the interval $[0,1]$. 
This experimental design is the same as that of \citet{zhu2020fully}.
The dimension of $x$ is set to $d=5,20$.
We consider different combination of the learning rate $\gamma_{t}=\gamma_0 t^{-a}$ by setting $\gamma_0=0.5, 1$ and $a = 0.505, 0.667$. 
The sample size set to be $n=100000$. 
The initial value $\beta_0$ is set to be zero. 
In case of $d=20$, we burn in around 1\% of observations and start to estimate $\bar{\beta}_t$ from $t=1000$. 
Finally, the simulation results are based on $1000$ replications.

We compare the performance of the proposed random scaling method with the state-of-the-art methods in the literature, especially the plug-in method in \citet{chen2020statistical} and the recursive batch-mean method in \citet{zhu2020fully}.
 The performance is measured by three statistics: the coverage rate, the average length of the 95\% confidence interval, and the average computation time. 
Note that the nominal coverage probability is set at 0.95. 
For brevity, we focus on the first coefficient $\beta_1$ hereafter. The results are similar across different coefficients.






\begin{figure}[htp]
\caption{Linear Regression: $d \in \{5, 20\}$, $\gamma_0=0.5$, $a= 0.505$  for $\gamma_t=\gamma_0 t^{-a}$}\label{fig:m03-maintext}
\centering
\begin{tabular}{c}
\multicolumn{1}{c}{\underline{$d=5$}}\\
\includegraphics[scale=0.15]{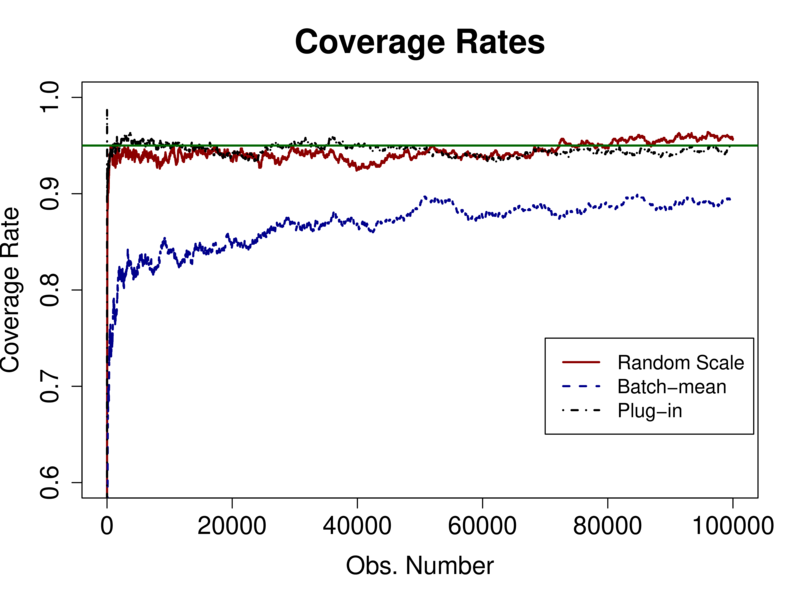}
\includegraphics[scale=0.15]{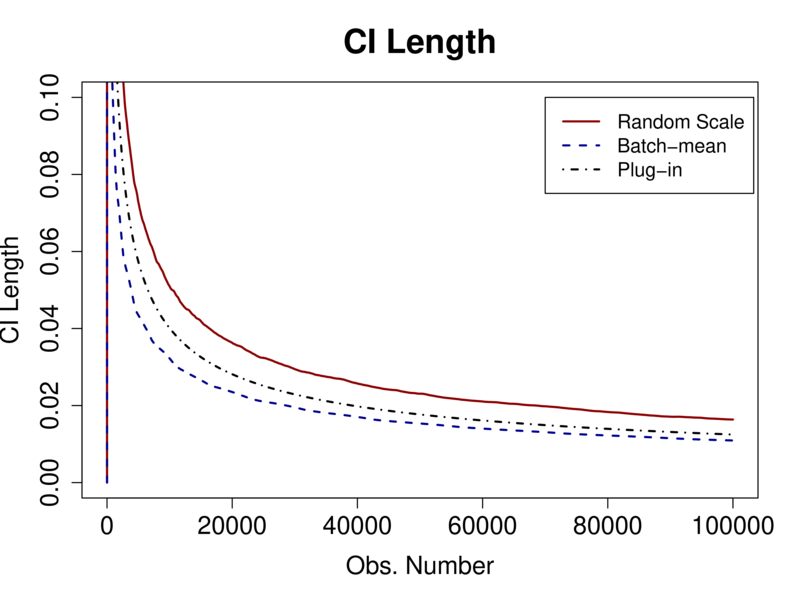}
\includegraphics[scale=0.15]{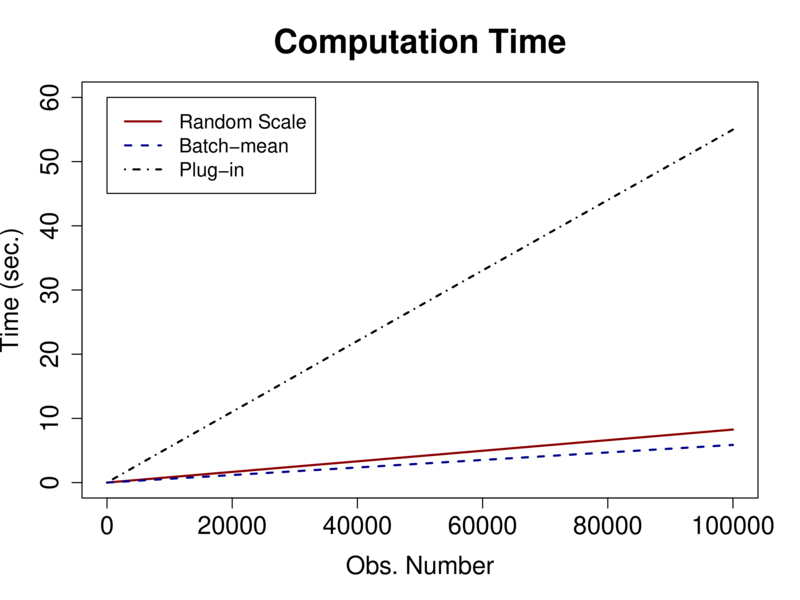}\\
\\
\multicolumn{1}{c}{\underline{$d=20$}}\\
\includegraphics[scale=0.15]{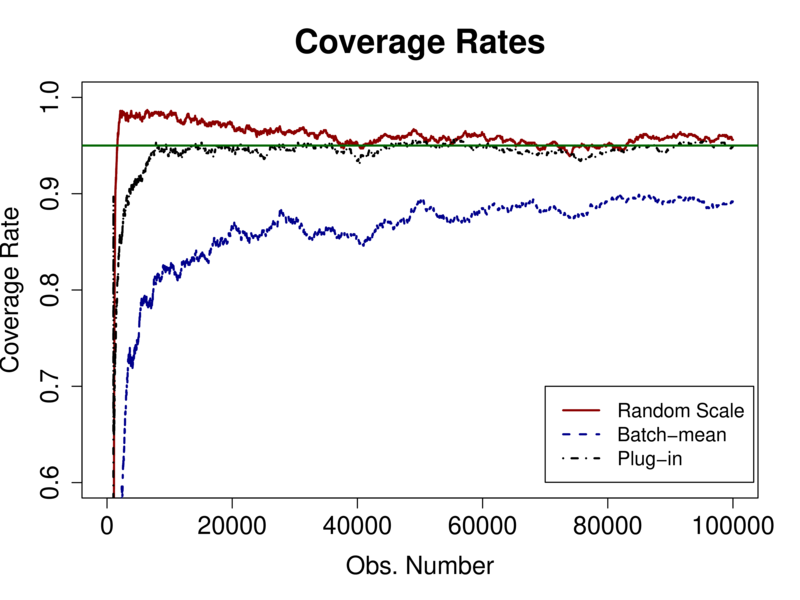}
\includegraphics[scale=0.15]{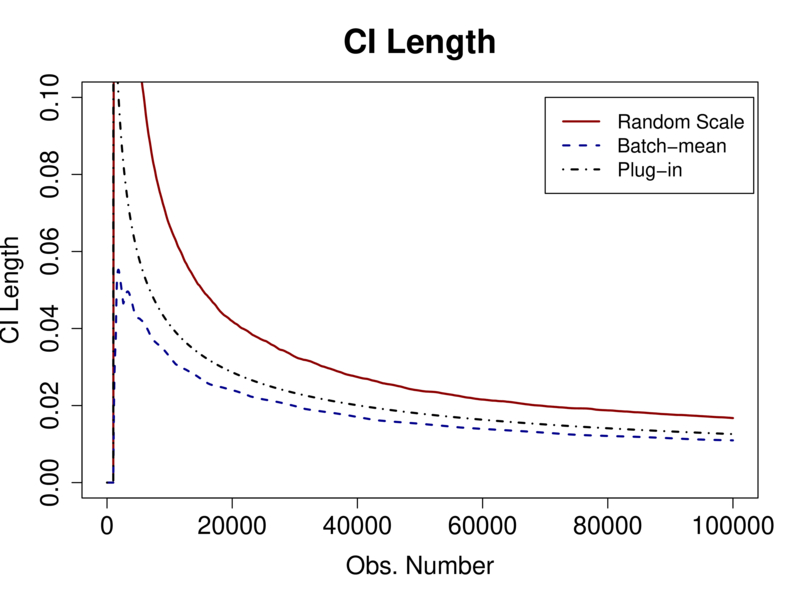}
\includegraphics[scale=0.15]{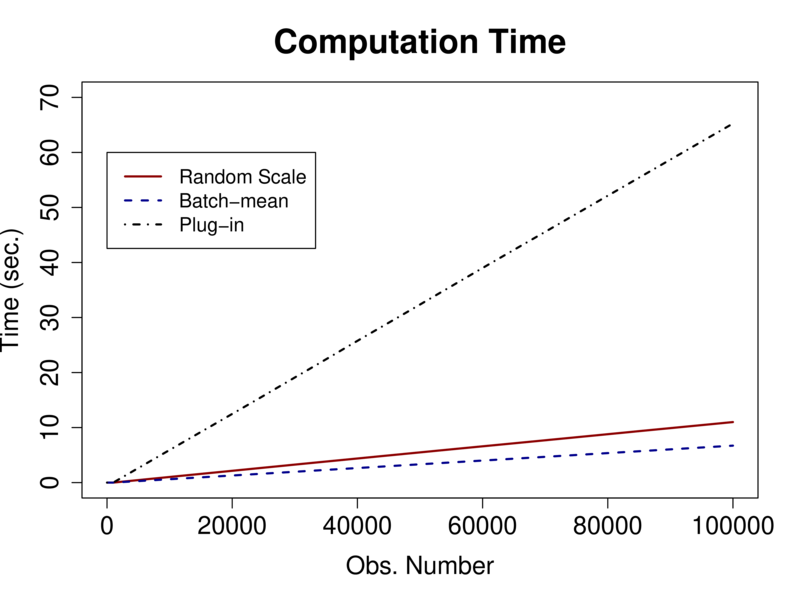}\\
\end{tabular}
\end{figure}

\begin{figure}[hpbt]
\caption{Linear Regression: $d = 5$, $\gamma_0\in\{0.5, 1\}$, $a\in\{0.505, 0.667\}$  for $\gamma_t=\gamma_0 t^{-a}$}\label{fig:compare-lr-maintext}
\centering
\begin{tabular}{cc}
{\underline{$\gamma_0=0.5, a=0.505$}} & {\underline{$\gamma_0=0.5, a=0.667$}} \\
\includegraphics[scale=0.20]{figures/fig-coverage-d5-m01.png} & \includegraphics[scale=0.20]{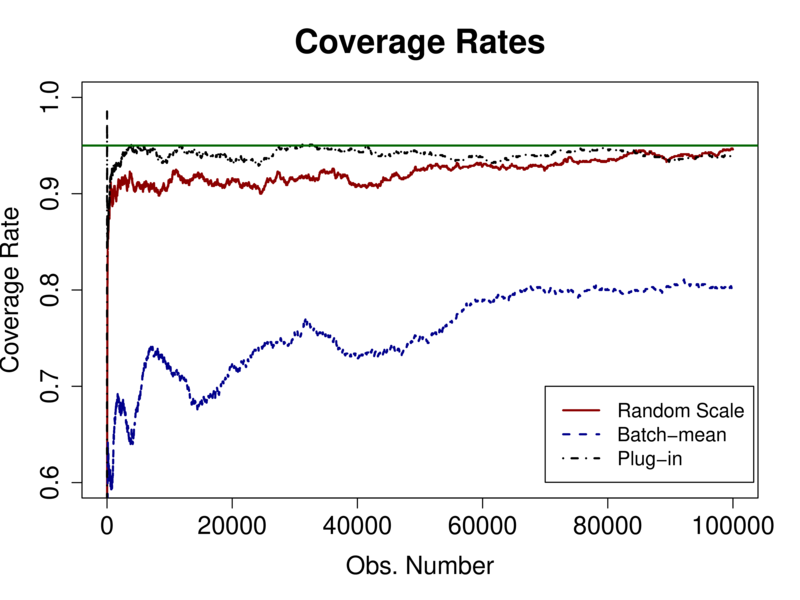} \\
{\underline{$\gamma_0=1, a=0.505$}} & {\underline{$\gamma_0=1, a=0.667$}}\\
 \includegraphics[scale=0.20]{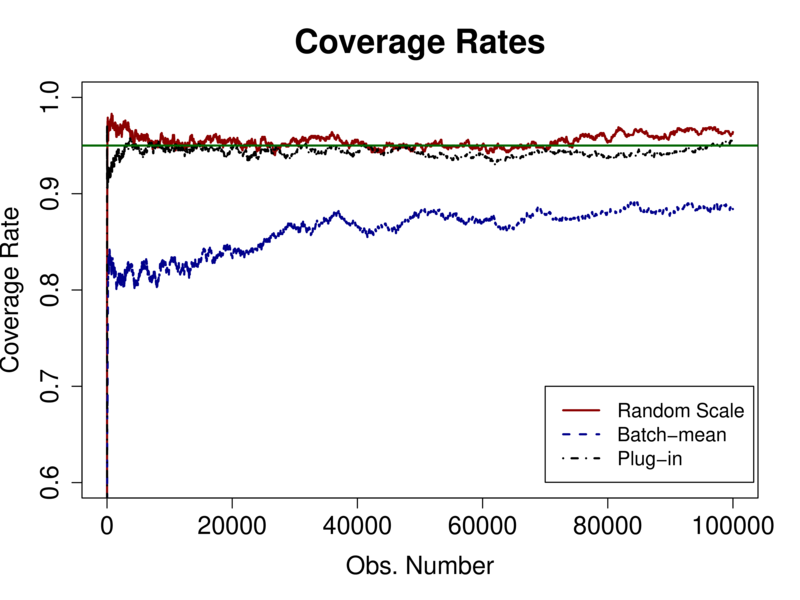} & \includegraphics[scale=0.20]{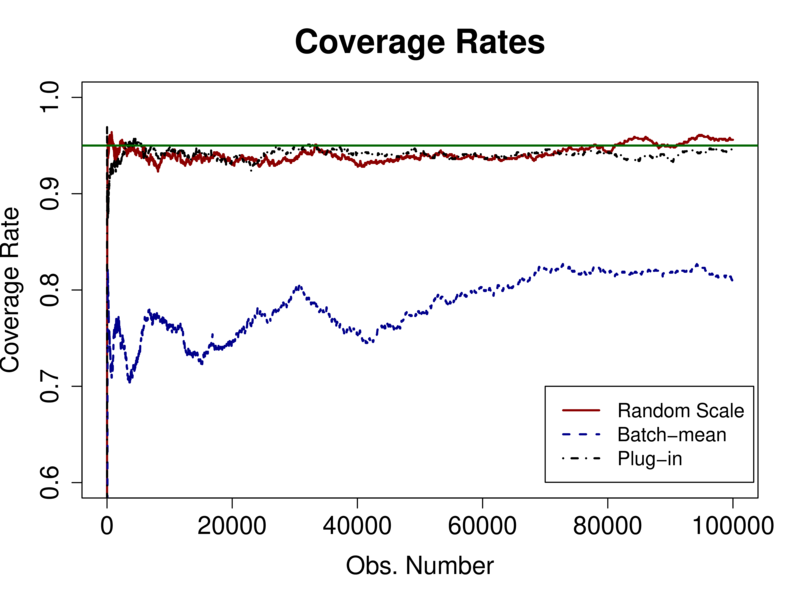}\\
\\
\end{tabular}
\end{figure}



Figures~\ref{fig:m03-maintext}--\ref{fig:compare-lr-maintext} summarize the simulation results. The complete set of simulation results are reported in the Appendix. 
In Figure~\ref{fig:m03-maintext}, we adopt the same learning rate parameters as in \citet{zhu2020fully}:   $\gamma_0=0.5$ and $a=0.505$.  
Overall, the performance of the random scaling method is satisfactory. 
First, the random scaling and plug-in methods show better coverage rates. 
The coverage rate of the batch-mean method deviates more than 5\% from the nominal rate even at $n=100000$.
Second, the batch-mean method shows the smallest average length of the confidence interval followed by the plug-in and the random scaling methods. 
Third, the plug-in method requires substantially more time for computation than the other two methods. 
The random scaling method takes slightly more computation time than the batch-mean method. 
Finally, we check the robustness of the performance by changing the learning rates in Figure~\ref{fig:compare-lr-maintext}, focusing on the case of $d=5$.
Both the random scaling method and the plug-in method are robust to the changes in the learning rates in terms of the coverage rate.
However, the batch-mean method converges slowly when $a=0.667$ and it deviates from the nominal rates about 15\% even at $n=100000$.

\paragraph{Logistic Regression}
We next turn our attention to the following logistic regression model:
\begin{align*}
y_t = 1(x_t'\beta^* - \varepsilon_t \ge 0)~~\mbox{for}~~t=1,\ldots,n,
\end{align*}
where $\varepsilon_t$ follows the standard logistic distribution and $1(\cdot)$ is the indicator function. 
We consider a large dimension of \(x_t\) (\(d=200\)) as well as \(d=5, 20\). All other settings are the same as the linear model.

\begin{table}[htbp]
\centering
\caption{Logistic Regression, $n=10^5$, $\gamma_0=0.5$, $a=0.505$  for $\gamma_t=\gamma_0 t^{-a}$} \label{tb:logit}
\begin{tabular}{lccc}
\hline
                        &    $d=5$     &     $d=20$     &   $d=200$        \\
\hline
 \underline{Random Scale}      \\ 
Coverage               &    0.930     &     0.929      &     0.919        \\
Length                 &    0.036     &     0.043      &     0.066   \\     
Time (sec.)            &     8.4      &      11.4      &     170.3     \\  
\hline
\underline{Batch-mean}  \\
Coverage               &    0.824 &     0.772      &     0.644        \\
Length                 &    0.022     &     0.024      &     0.027       \\ 
Time (sec.)            &     6.0      &       7.0      &      10.7       \\
\hline
\underline{Plug-in}       \\
Coverage              &    0.953     &     0.946      &     0.944        \\
Length                 &    0.029     &     0.035      &     0.053        \\
Time (sec.)            &    55.2      &      66.8      &     955.0       \\
\hline

\end{tabular}
\end{table}

Overall, the simulation results are similar to those in linear regression. 
Table \ref{tb:logit} summarizes the simulation results of a single design.
The coverage rates of Random Scale and Plug-in are satisfactory while that of Batch-mean is 30\% lower when $d=200$.
Random Scale requires more computation time than Batch-mean but is still much faster than Plug-in. 
The computation time of Random Scale can be substantially reduced when we are interested in the inference of a single parameter. In such a case, we need to update only a single element of $\hat{V}$ rather than the whole $d\times d$ matrix. 
In Table \ref{tb:logit_single}, we show that Random Scale can be easily scaled up to $d=800$ with only 11.7 seconds computation time when we are interested in the inference of a single parameter. 
Finally, the results in the appendix reinforce our findings from the linear regression design that the performance of Random-scale is less sensitive to the choice of tuning parameters than Batch-mean.

\begin{table}[htbp]
\centering
\caption{Logistic Regression: Random Scale Updating a Single Element of $\hat{V}$, $n=10^5$, $\gamma_0=0.5$, $a=0.505$  for $\gamma_t=\gamma_0 t^{-a}$} \label{tb:logit_single}
\begin{tabular}{lccccc}
\hline
& $d=5$ & $d=20$ & $d=200$ & $d=500$ & $d=800$\\
\hline
Coverage &  0.930 & 0.929 & 0.919 & 0.927 & 0.931\\
Length  & 0.037     & 0.043 &   0.066 & 0.133 & 0.196\\
Time (sec.) &   5.0 &   5.3 &   6.7 &   9.7 &   11.7\\
\hline
\end{tabular}
\end{table}

\clearpage
\appendix

\section*{Appendix: Additional Experiment Results}\label{sec:additional_results}


In this section we provide the complete set of simulation results. The R codes to replicate the results are also included in this supplemental package. 
\bigskip 

\noindent \textbf{Linear Regression: } Recall that we the linear regression model is generated from
\begin{align*}
y_{t} = x_t'\beta^* + \varepsilon_t~~\mbox{for}~~t=1,\ldots,n,
\end{align*}
where $x_t$ is a $d$-dimensional covariates generated from the multivariate normal distribution $\mathcal{N} (0,I_d)$, $\varepsilon_t$ is from $N(0,1)$, and $\beta^*$ is equi-spaced on the interval $[0,1]$. 
The dimension of $x$ is set to $d=5,20$.
We consider different combination of the learning rate $\gamma_{t}=\gamma_0 t^{-a}$ with $1/2<a<1$, where $\gamma_0=0.5, 1$ and $a = 0.505, 0.667$. 
The sample size set to be $n=10^5$. 
The initial value $\beta_0$ is set to be zero. 
In case of $d=20$, we burn in around 1\% of observations and start to estimate $\bar{\beta}_t$ from $t=1000$. 
Finally, the simulation results are based on $1000$ replications.

\begin{figure}[htp]
\caption{Comparison over Different Coefficients: $d=5$} \label{sfig:diff-coef}
\centering
\begin{tabular}{c}
\multicolumn{1}{c}{\underline{$\gamma_0=0.5, a=0.505$}}\\
\includegraphics[scale=0.20]{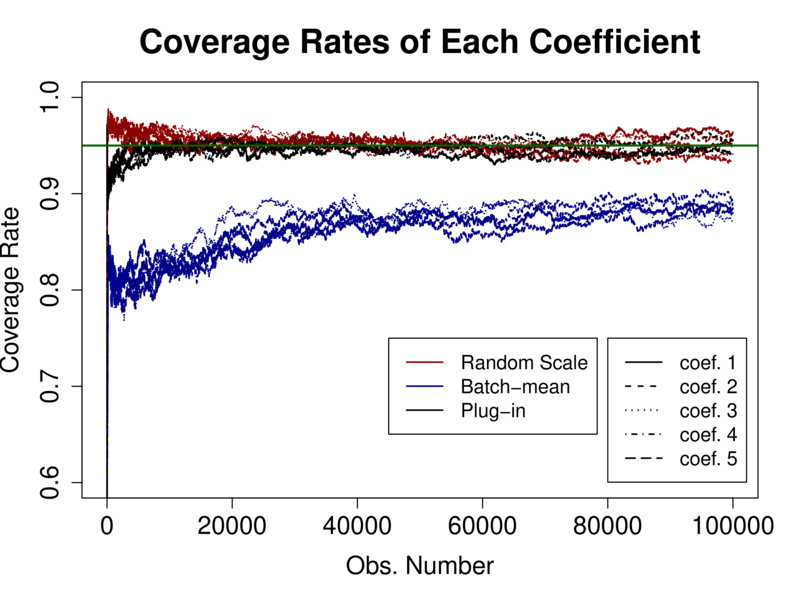}
\includegraphics[scale=0.20]{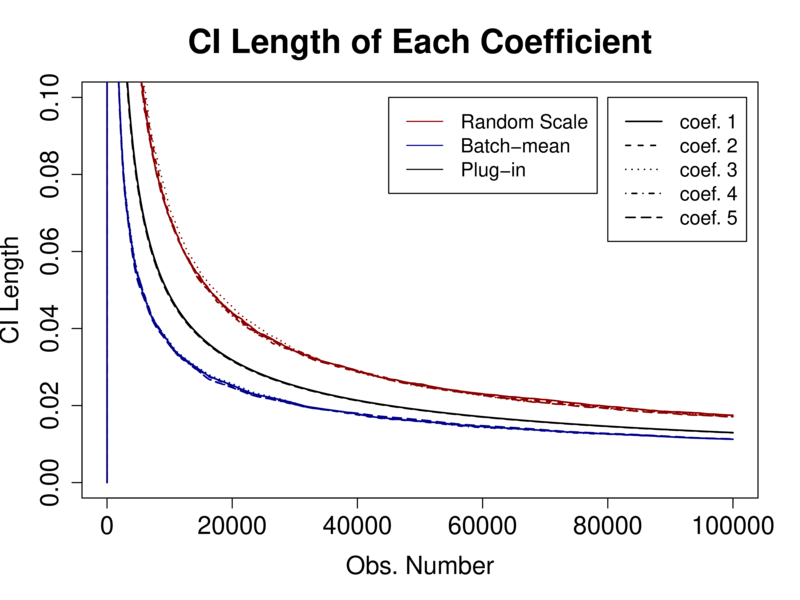}\\
\multicolumn{1}{c}{\underline{$\gamma_0=0.5, a=0.667$}}\\
\includegraphics[scale=0.20]{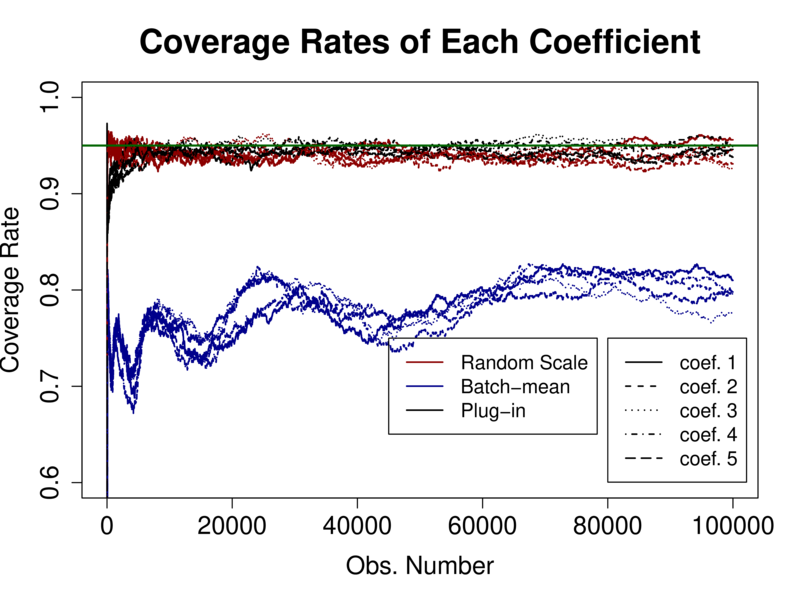}
\includegraphics[scale=0.20]{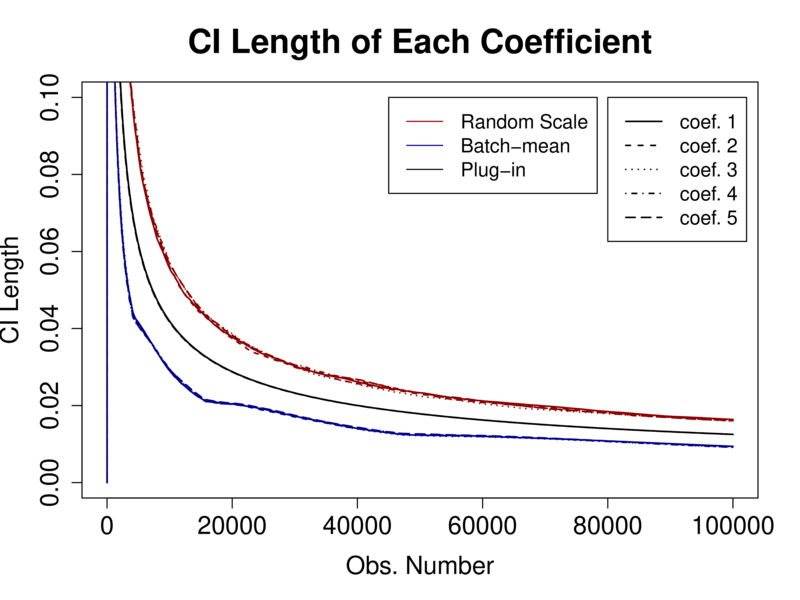}\\
\multicolumn{1}{c}{\underline{$\gamma_0=1, a=0.505$}}\\
\includegraphics[scale=0.20]{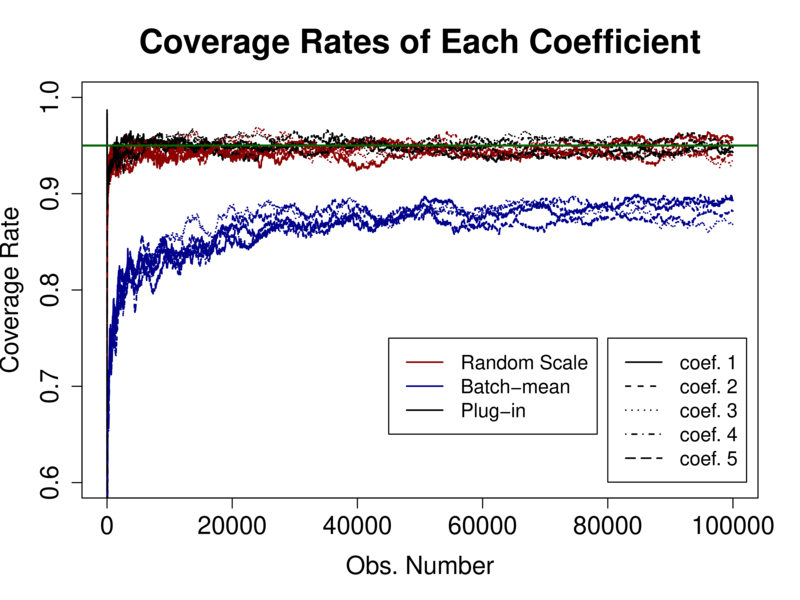}
\includegraphics[scale=0.20]{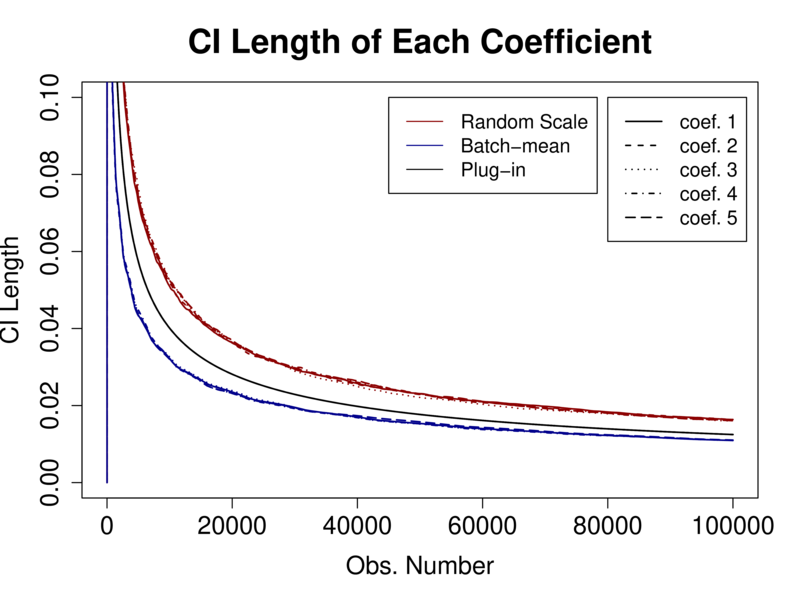}\\
\multicolumn{1}{c}{\underline{$\gamma_0=1, a=0.667$}}\\
\includegraphics[scale=0.20]{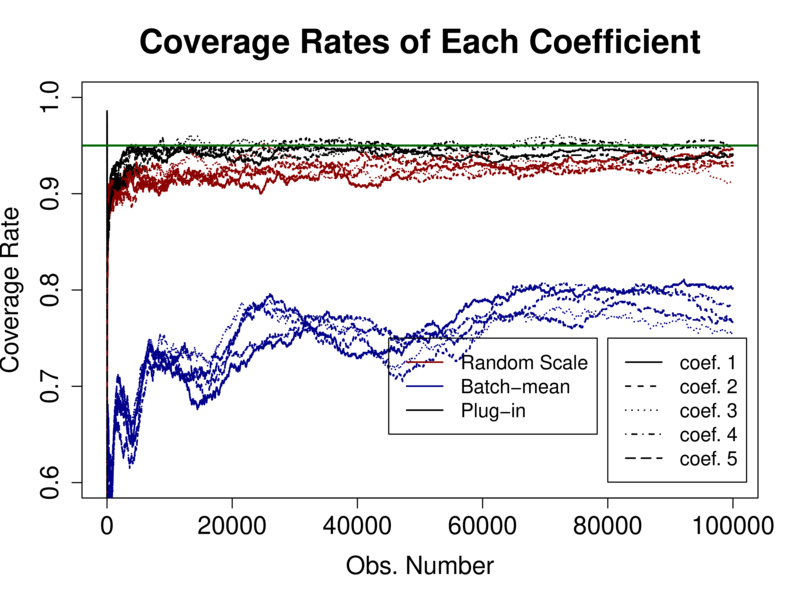}
\includegraphics[scale=0.20]{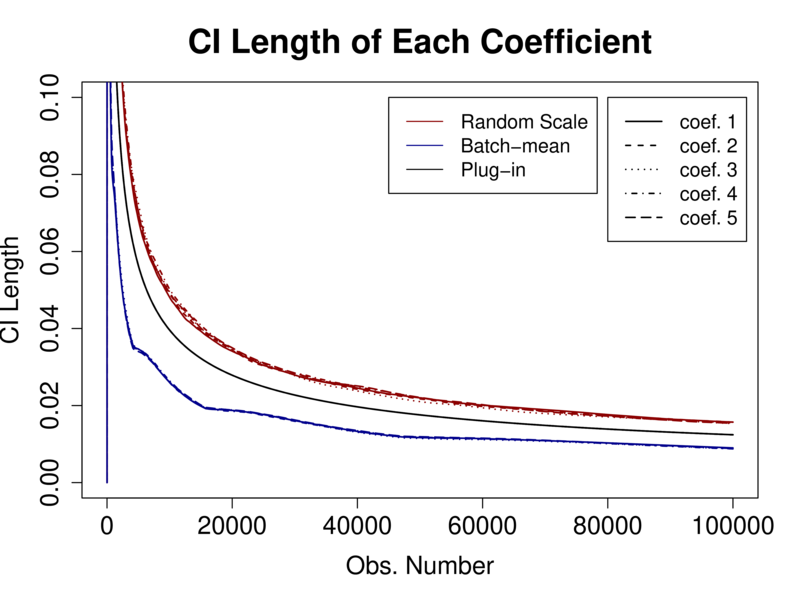}\\
\end{tabular}
\end{figure}

Figure~\ref{sfig:diff-coef} shows the coverage rates and the lengths of the confidence interval for all five coefficients when $d=5$. As discussed in the main text, we observe similar behavior for different coefficients over different learning rates. 

\begin{figure}[htp]
\caption{Linear Model: $d=5$, $\gamma_0\in\{0.5, 1\}$, $a\in\{0.505, 0.667\}$   for $\gamma_t=\gamma_0 t^{-a}$} \label{sfig:d5}
\centering
\begin{tabular}{c}
\multicolumn{1}{c}{\underline{$\gamma_0=0.5, a=0.505$}}\\
\includegraphics[scale=0.15]{figures/fig-coverage-d5-m01.png}
\includegraphics[scale=0.15]{figures/fig-length-d5-m01.png}
\includegraphics[scale=0.15]{figures/fig-time-d5-m01.png}\\
\multicolumn{1}{c}{\underline{$\gamma_0=0.5, a=0.667$}}\\
\includegraphics[scale=0.15]{figures/fig-coverage-d5-m02.png}
\includegraphics[scale=0.15]{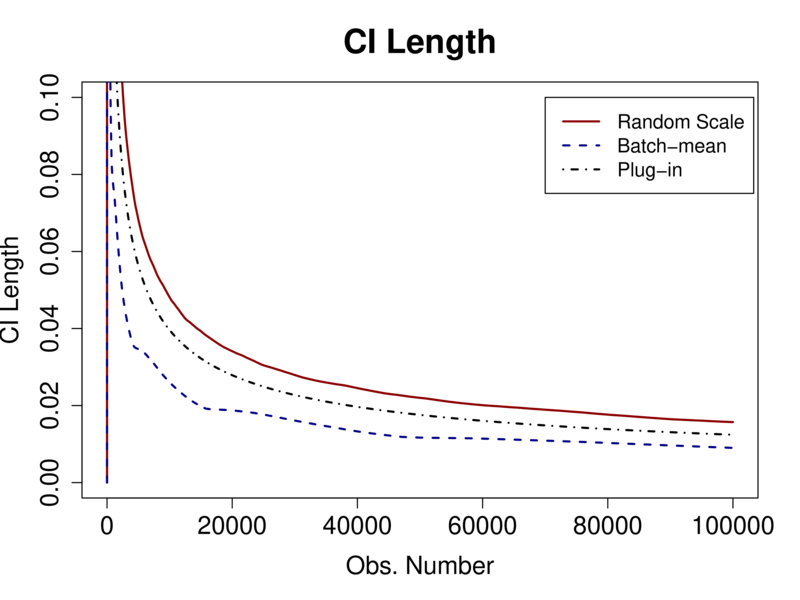}
\includegraphics[scale=0.15]{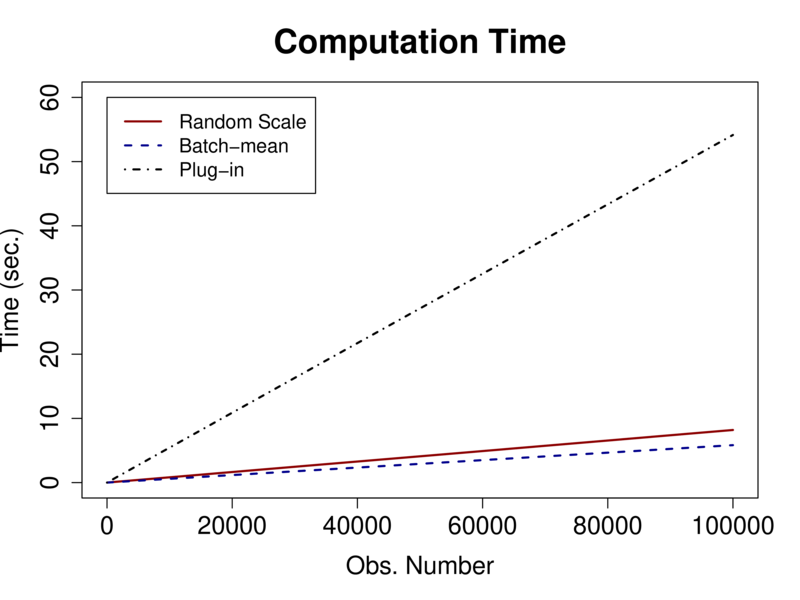}\\
\multicolumn{1}{c}{\underline{$\gamma_0=1, a=0.505$}}\\
\includegraphics[scale=0.15]{figures/fig-coverage-d5-m03.png}
\includegraphics[scale=0.15]{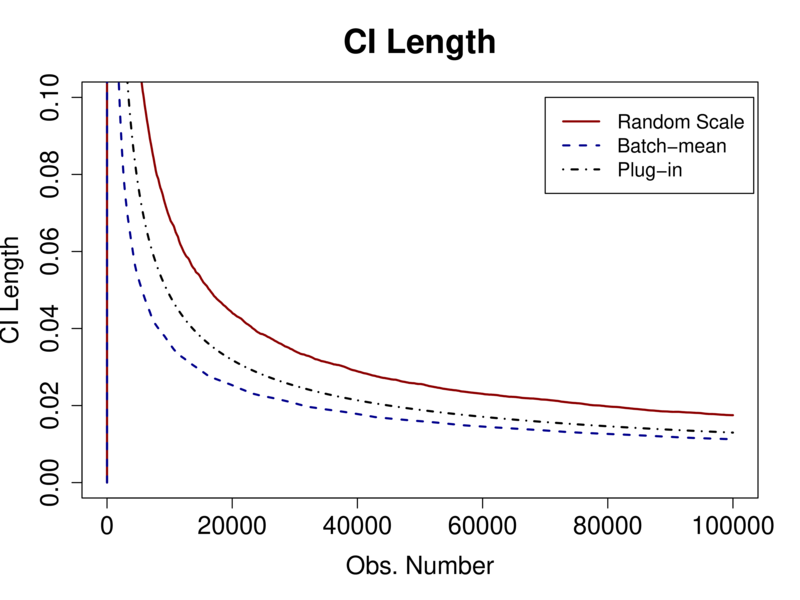}
\includegraphics[scale=0.15]{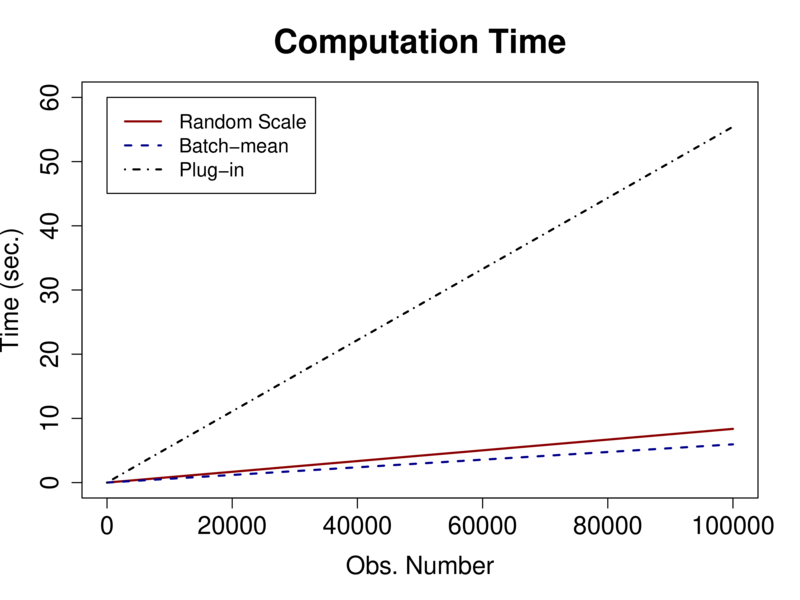}\\
\multicolumn{1}{c}{\underline{$\gamma_0=1, a=0.667$}}\\
\includegraphics[scale=0.15]{figures/fig-coverage-d5-m04.png}
\includegraphics[scale=0.15]{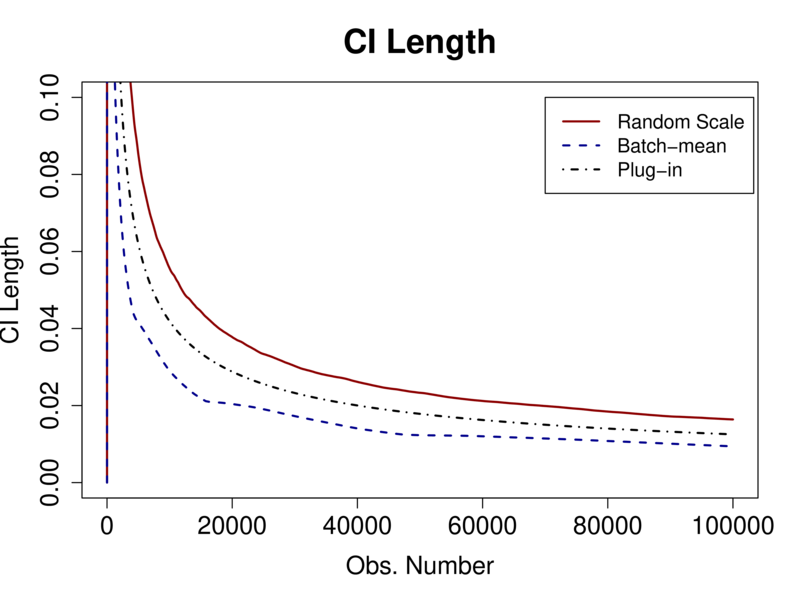}
\includegraphics[scale=0.15]{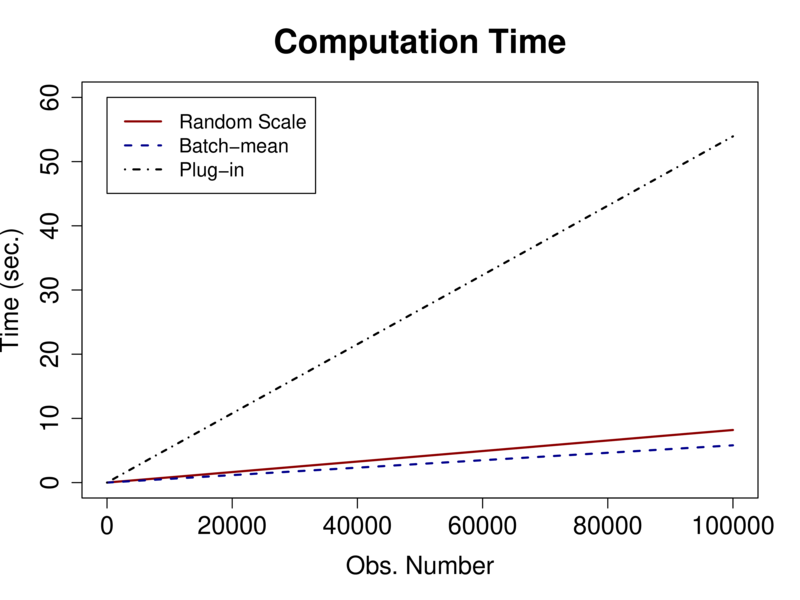}\\
\end{tabular}
\end{figure}

\begin{figure}[htp]
\caption{Linear Model: $d=20$, $\gamma_0\in\{0.5, 1\}$, $a\in\{0.505, 0.667\}$  for $\gamma_t=\gamma_0 t^{-a}$}\label{sfig:d20}
\centering
\begin{tabular}{c}
\\
\multicolumn{1}{c}{\underline{$\gamma_0=0.5, a=0.505$}}\\
\includegraphics[scale=0.15]{figures/fig-coverage-d20-m01.png}
\includegraphics[scale=0.15]{figures/fig-length-d20-m01.png}
\includegraphics[scale=0.15]{figures/fig-time-d20-m01.png}\\
\\
\multicolumn{1}{c}{\underline{$\gamma_0=0.5, a=0.667$}}\\
\includegraphics[scale=0.15]{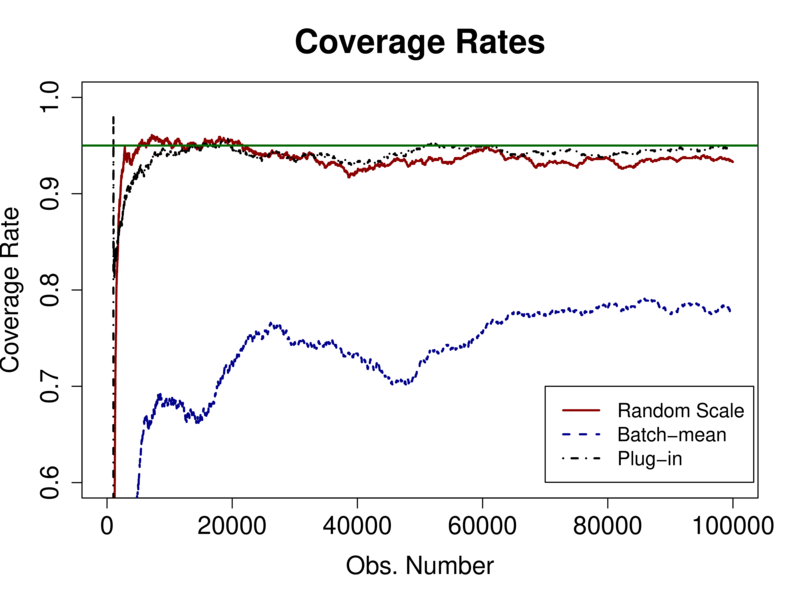}
\includegraphics[scale=0.15]{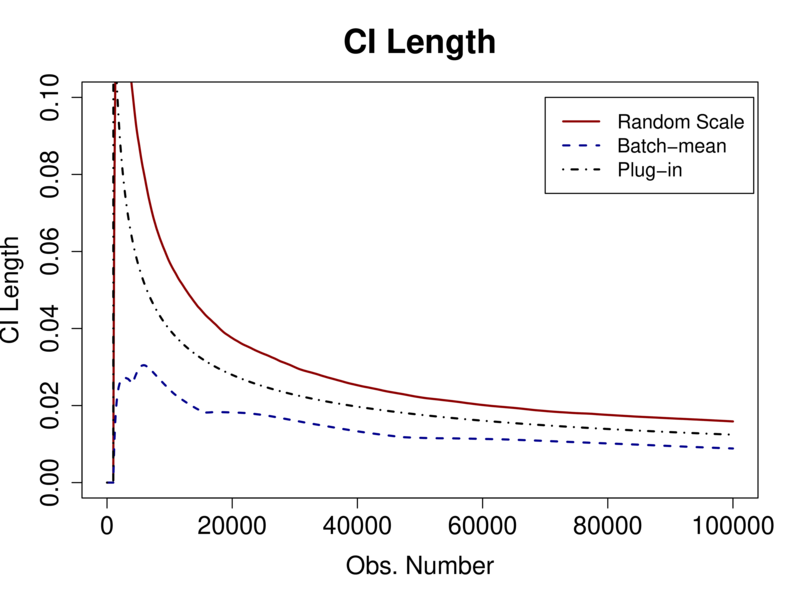}
\includegraphics[scale=0.15]{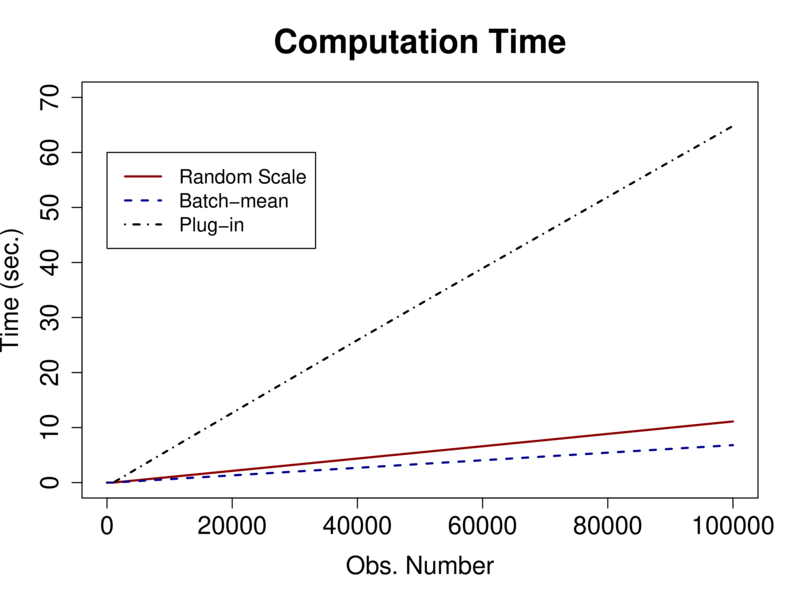}\\
\\
\multicolumn{1}{c}{\underline{$\gamma_0=1, a=0.505$}}\\
\includegraphics[scale=0.15]{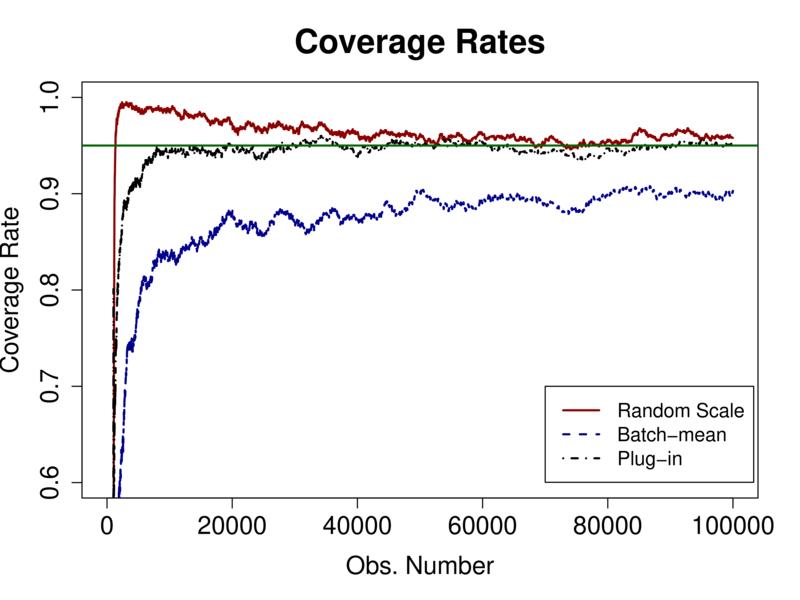}
\includegraphics[scale=0.15]{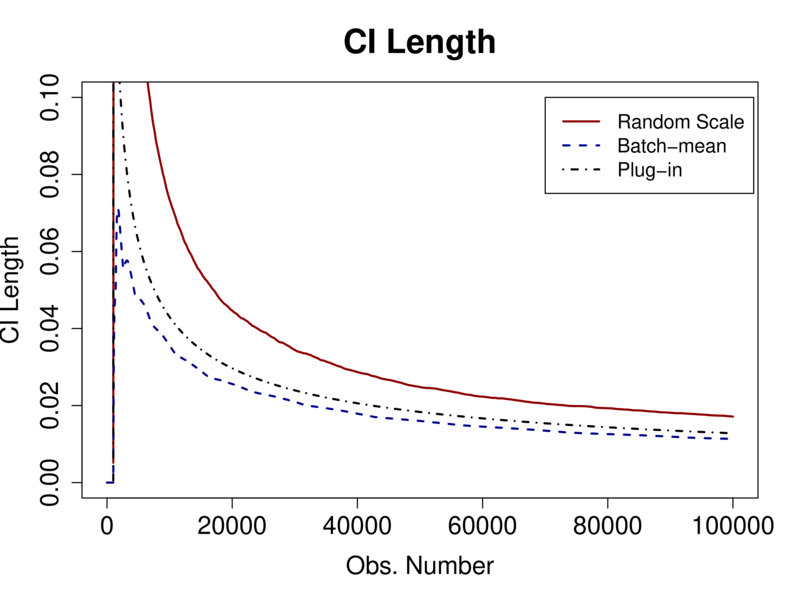}
\includegraphics[scale=0.15]{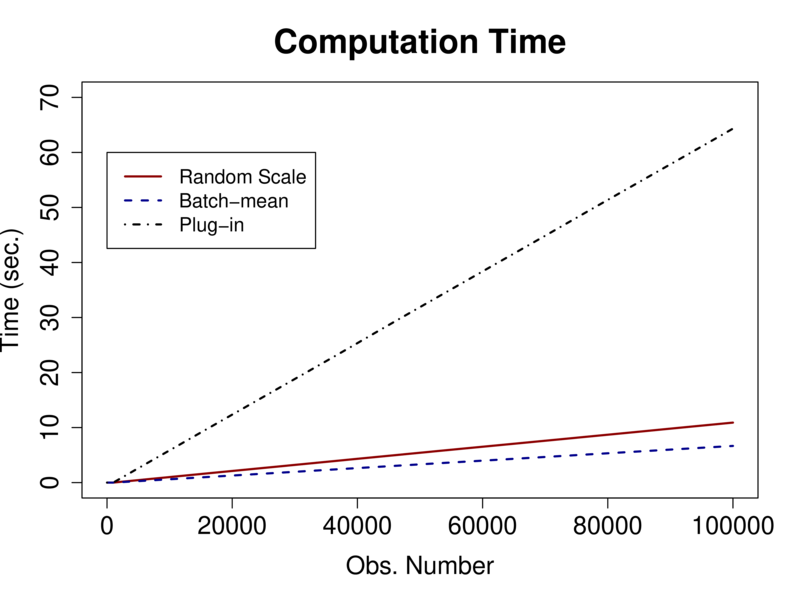}\\
\\
\multicolumn{1}{c}{\underline{$\gamma_0=1, a=0.667$}}\\
\includegraphics[scale=0.15]{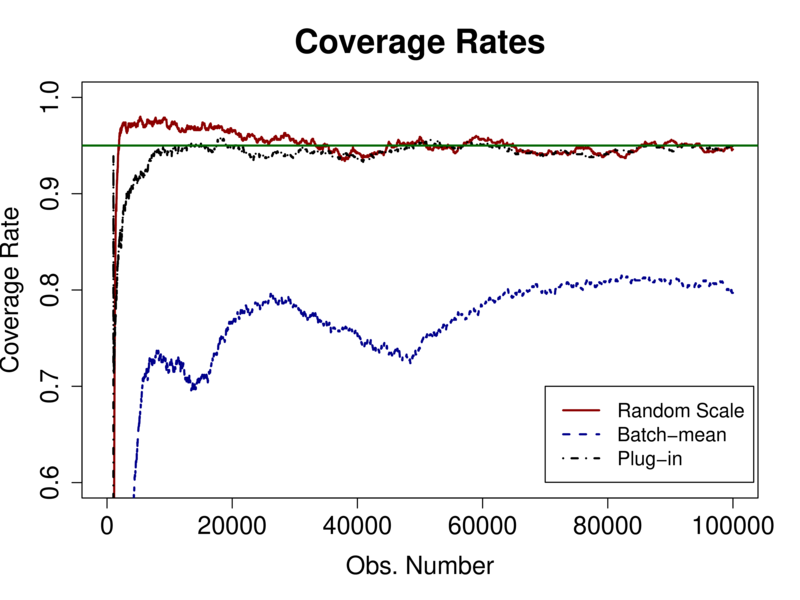}
\includegraphics[scale=0.15]{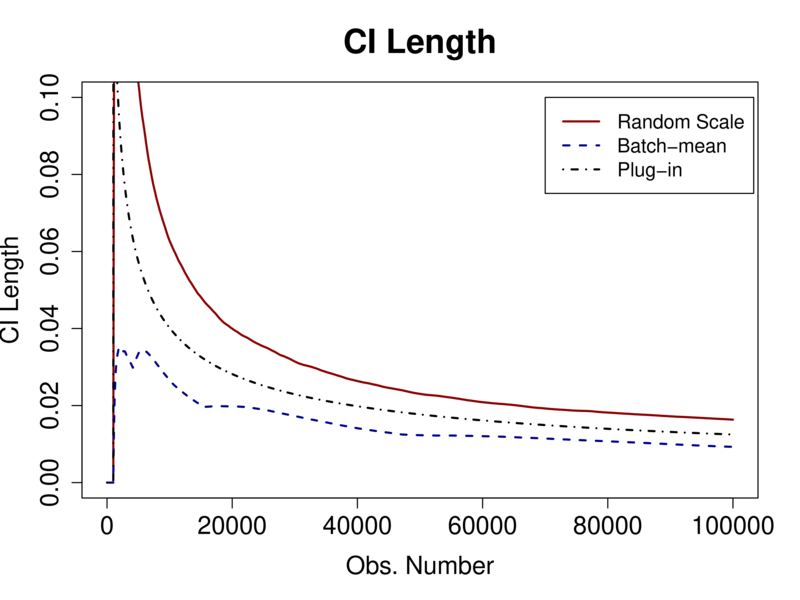}
\includegraphics[scale=0.15]{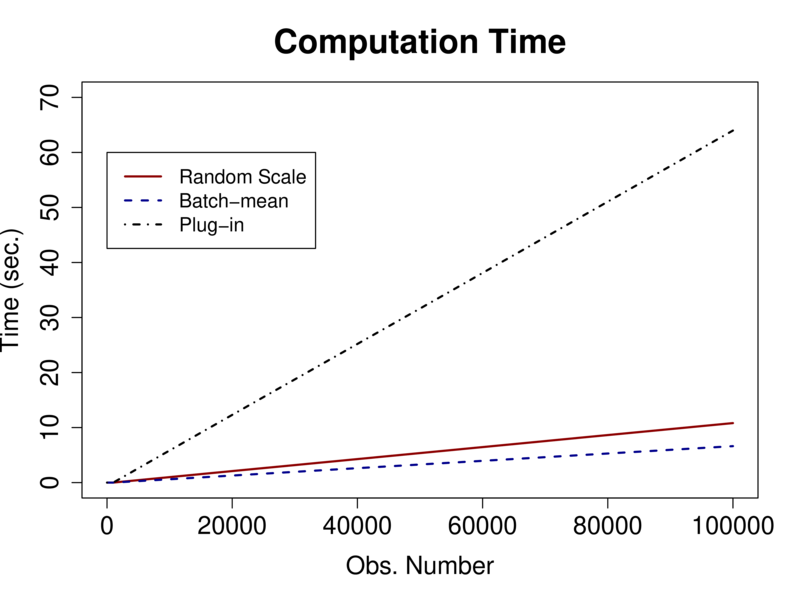}\\
\end{tabular}
\end{figure}

Figures~\ref{sfig:d5}--\ref{sfig:d20} show the complete paths of coverage rates, confidence interval lengths, and the computation times for each design with different learning rates. Tables~\ref{stb:d=5}--\ref{stb:d=20} report the same statistics at $n=25000, 50000, 75000,$ and $100000$. Overall, the results are in line with our discussion in the main text. The proposed random scaling method and the plug-in method fit the size well while the batch-mean method show less precise coverage rates. The computation time of the plug-in method is at least five times slower than the other two methods. Because the batch size depends on the learning rate parameter $a$, the batch-mean results are sensitive to different learning rates.

\begin{table}[ht]
\centering
\footnotesize
\caption{Linear Model: $d=5$, $\gamma_0\in\{0.5, 1\}$, $a\in\{0.505, 0.667\}$  for $\gamma_t=\gamma_0 t^{-a}$}
\label{stb:d=5}
\begin{tabular}{llcccc}
  \toprule
  & & $n=25000$ & $n=50000$ & $n=75000$ & $n=100000$ \\
  \midrule
  \multicolumn{3}{l}{\underline{$\gamma_0=0.5, a=0.505$}}\\
  \multirow{3}{*}{Coverage} 
  & Random Scale & 0.941 (0.0075) & 0.942 (0.0074) & 0.950 (0.0069) & 0.957 (0.0064) \\ 
  & Batch-mean   & 0.859 (0.0110) & 0.888 (0.0100) & 0.883 (0.0102) & 0.893 (0.0098) \\ 
  & Plug-in      & 0.947 (0.0071) & 0.947 (0.0071) & 0.947 (0.0071) & 0.948 (0.0070) \\ 
  \hline
  \multirow{3}{*}{Length} 
  & Random Scale & 0.032 & 0.023 & 0.019 & 0.016 \\ 
  & Batch-mean   & 0.021 & 0.015 & 0.013 & 0.011 \\ 
  & Plug-in      & 0.025 & 0.018 & 0.014 & 0.012 \\ 
  \hline
  \multirow{3}{*}{Time (sec.)} 
  & Random Scale & 2.1 & 4.1 & 6.2 & 8.3 \\ 
  & Batch-mean   & 1.5 & 2.9 & 4.4 & 5.9 \\ 
  & Plug-in      & 13.8 & 27.6 & 41.3 & 55.0 \\ 
  \hline
  \\
  \multicolumn{3}{l}{\underline{$\gamma_0=0.5, a=0.667$}}\\
  \multirow{3}{*}{Coverage} 
  & Random Scale & 0.905 (0.0093) & 0.924 (0.0084) & 0.933 (0.0079) & 0.946 (0.0071) \\ 
  & Batch-mean   & 0.741 (0.0139) & 0.752 (0.0137) & 0.793 (0.0128) & 0.802 (0.0126) \\ 
  & Plug-in      & 0.934 (0.0079) & 0.942 (0.0074) & 0.945 (0.0072) & 0.940 (0.0075) \\ 
  \hline
  \multirow{3}{*}{Length} 
  & Random Scale & 0.030  & 0.022  & 0.018 & 0.016 \\ 
  & Batch-mean   & 0.018  & 0.012  & 0.011 & 0.009 \\ 
  & Plug-in      & 0.025  & 0.018  & 0.014 & 0.012 \\ 
  \hline
  \multirow{3}{*}{Time (sec.)} 
  & Random Scale & 2.1 & 4.1 & 6.1 & 8.2 \\ 
  & Batch-mean   & 1.5 & 2.9 & 4.4 & 5.8 \\ 
  & Plug-in      & 13.6 & 27.1 & 40.6 & 54.1 \\ 
  \hline
  \\
  \multicolumn{3}{l}{\underline{$\gamma_0=1, a=0.505$}}\\
  \multirow{3}{*}{Coverage} 
  & Random Scale & 0.957 (0.0064) & 0.950 (0.0069) & 0.958 (0.0063) & 0.964 (0.0059) \\ 
  & Batch-mean   & 0.849 (0.0113) & 0.880 (0.0103) & 0.875 (0.0105) & 0.884 (0.0101) \\ 
  & Plug-in      & 0.947 (0.0071) & 0.941 (0.0075) & 0.943 (0.0073) & 0.956 (0.0065) \\
  \hline 
  \multirow{3}{*}{Length} 
  & Random Scale & 0.038 & 0.026 & 0.021  & 0.018  \\ 
  & Batch-mean   & 0.022 & 0.016 & 0.013  & 0.011  \\ 
  & Plug-in      & 0.028 & 0.019 & 0.015  & 0.013  \\ 
  \hline
  \multirow{3}{*}{Time (sec.)} 
  & Random Scale & 2.1 & 4.2 & 6.3 & 8.4 \\ 
  & Batch-mean   & 1.5 & 3.0 & 4.5 & 6.0 \\ 
  & Plug-in      & 13.9 & 27.7 & 41.6 & 55.4 \\ 
  \hline
   \\
  \multicolumn{3}{l}{\underline{$\gamma_0=1, a=0.667$}}\\
  \multirow{3}{*}{Coverage} 
  & Random Scale & 0.933 (0.0079) & 0.936 (0.0077) & 0.946 (0.0071) & 0.956 (0.0065) \\ 
  & Batch-mean   & 0.773 (0.0132) & 0.774 (0.0132) & 0.820 (0.0121) & 0.810 (0.0124) \\ 
  & Plug-in      & 0.942 (0.0074) & 0.942 (0.0074) & 0.943 (0.0073) & 0.946 (0.0071) \\ 
  \hline 
  \multirow{3}{*}{Length} 
  & Random Scale & 0.033 & 0.023 & 0.019 & 0.016 \\ 
  & Batch-mean   & 0.019 & 0.012 & 0.011 & 0.009 \\ 
  & Plug-in      & 0.026 & 0.018 & 0.014 & 0.013 \\ 
  \hline
  \multirow{3}{*}{Time (sec.)} 
  & Random Scale & 2.1 & 4.1 & 6.1 & 8.2 \\ 
  & Batch-mean   & 1.5 & 2.9 & 4.4 & 5.8 \\ 
  & Plug-in      & 13.5 & 26.9 & 40.4 & 53.9 \\ 
  \bottomrule
\end{tabular}
\end{table}

\begin{table}[ht]
\centering
\footnotesize
\caption{Linear Model: $d=20$, $\gamma_0\in\{0.5, 1\}$, $a\in\{0.505, 0.667\}$  for $\gamma_t=\gamma_0 t^{-a}$}
\label{stb:d=20}
\begin{tabular}{llcccc}
  \toprule
  & & $n=25000$ & $n=50000$ & $n=75000$ & $n=100000$ \\
 \midrule
  \multicolumn{3}{l}{\underline{$\gamma_0=0.5, a=0.505$}}\\
  \multirow{3}{*}{Coverage} 
  & Random Scale & 0.963 (0.0060) & 0.960 (0.0062) & 0.949 (0.0070) & 0.956 (0.0065) \\ 
  & Batch-mean   & 0.860 (0.0110) & 0.890 (0.0099) & 0.878 (0.0103) & 0.890 (0.0099) \\ 
  & Plug-in      & 0.938 (0.0076) & 0.954 (0.0066) & 0.937 (0.0077) & 0.949 (0.0070) \\ 
  \hline
  \multirow{3}{*}{Length} 
  & Random Scale & 0.037 & 0.024 & 0.019 & 0.017 \\ 
  & Batch-mean   & 0.022 & 0.015 & 0.012 & 0.011 \\ 
  & Plug-in      & 0.026 & 0.018 & 0.015 & 0.013 \\ 
  \hline
  \multirow{3}{*}{Time (sec.)} 
  & Random Scale & 2.7 & 5.5 & 8.2 & 11.0 \\ 
  & Batch-mean   & 1.6 & 3.3 & 5.0 & 6.7 \\ 
  & Plug-in      & 15.8 & 32.4 & 48.9 & 65.2 \\ 
  \hline
  \\
  \multicolumn{3}{l}{\underline{$\gamma_0=0.5, a=0.667$}}\\
  \multirow{3}{*}{Coverage} 
  & Random Scale & 0.941 (0.0075) & 0.935 (0.0078) & 0.928 (0.0082) & 0.933 (0.0079) \\ 
  & Batch-mean   & 0.751 (0.0137) & 0.724 (0.0141) & 0.778 (0.0131) & 0.777 (0.0132) \\ 
  & Plug-in      & 0.938 (0.0076) & 0.946 (0.0071) & 0.944 (0.0073) & 0.948 (0.0070) \\ 
  \hline
  \multirow{3}{*}{Length} 
  & Random Scale & 0.033 & 0.022 & 0.018 & 0.016 \\ 
  & Batch-mean   & 0.018 & 0.012 & 0.010 & 0.009 \\ 
  & Plug-in      & 0.025 & 0.018 & 0.014 & 0.012 \\ 
  \hline
  \multirow{3}{*}{Time (sec.)} 
  & Random Scale & 2.7 & 5.5 & 8.3 & 11.1 \\ 
  & Batch-mean   & 1.7 & 3.4 & 5.1 & 6.8 \\ 
  & Plug-in      & 16.0 & 32.4 & 48.7 & 64.8 \\ 
  \hline
  \\
  \multicolumn{3}{l}{\underline{$\gamma_0=1, a=0.505$}}\\
  \multirow{3}{*}{Coverage} 
  & Random Scale & 0.968 (0.0056) & 0.961 (0.0061) & 0.952 (0.0068) & 0.960 (0.0062) \\ 
  & Batch-mean   & 0.857 (0.0111) & 0.899 (0.0095) & 0.876 (0.0104) & 0.897 (0.0096) \\ 
  & Plug-in      & 0.942 (0.0074) & 0.952 (0.0068) & 0.938 (0.0076) & 0.954 (0.0066) \\ 
  \hline 
  \multirow{3}{*}{Length} 
  & Random Scale & 14.426 & 7.230 & 4.826 & 3.622 \\ 
  & Batch-mean   & 3.389 & 1.457 & 0.938 & 0.661 \\ 
  & Plug-in      & 16.776 & 8.359 & 5.549 & 4.151 \\ 
  \hline
  \multirow{3}{*}{Time (sec.)} 
  & Random Scale & 2.7 & 5.4 & 8.1 & 10.9 \\ 
  & Batch-mean   & 1.6 & 3.3 & 5.0 & 6.6 \\ 
  & Plug-in      & 15.5 & 31.7 & 47.8 & 63.9 \\ 
  \hline
   \\
  \multicolumn{3}{l}{\underline{$\gamma_0=1, a=0.667$}}\\
  \multirow{3}{*}{Coverage} 
  & Random Scale & 0.955 (0.0066) & 0.959 (0.0063) & 0.948 (0.0070) & 0.946 (0.0071) \\ 
  & Batch-mean   & 0.789 (0.0129) & 0.739 (0.0139) & 0.806 (0.0125) & 0.797 (0.0127) \\ 
  & Plug-in      & 0.939 (0.0076) & 0.953 (0.0067) & 0.943 (0.0073) & 0.949 (0.0070) \\ 
  \hline 
  \multirow{3}{*}{Length} 
  & Random Scale & 0.035 & 0.023 & 0.019 & 0.016 \\ 
  & Batch-mean   & 0.019 & 0.012 & 0.011 & 0.009 \\ 
  & Plug-in      & 0.025 & 0.018 & 0.014 & 0.012 \\ 
  \hline
  \multirow{3}{*}{Time (sec.)} 
  & Random Scale & 2.6 & 5.4 & 8.1 & 10.8 \\ 
  & Batch-mean   & 1.6 & 3.3 & 5.0 & 6.6 \\ 
  & Plug-in      & 15.5 & 31.6 & 47.8 & 64.0 \\ 
  \bottomrule
\end{tabular}
\end{table}

\clearpage

\noindent \textbf{Logistic Regression: } Recall that we the logistic regression model is generated from
\begin{align*}
y_{t} = 1(x_t'\beta^* - \varepsilon_t \ge 0 )~~\mbox{for}~~t=1,\ldots,n,
\end{align*}
where $\varepsilon_t$ follows the standard logistic distribution and $1(\cdot)$ is the indicator function. We consider $d=5,20, 200$. All other settings are the same as the linear model.

\begin{figure}[htp]
\caption{Logistic Model: $d=5$, $\gamma_0\in\{0.5, 1\}$, $a\in\{0.505, 0.667\}$   for $\gamma_t=\gamma_0 t^{-a}$} \label{fig_logit:d5}
\centering
\begin{tabular}{c}
\multicolumn{1}{c}{\underline{$\gamma_0=0.5, a=0.505$}}\\
\includegraphics[scale=0.15]{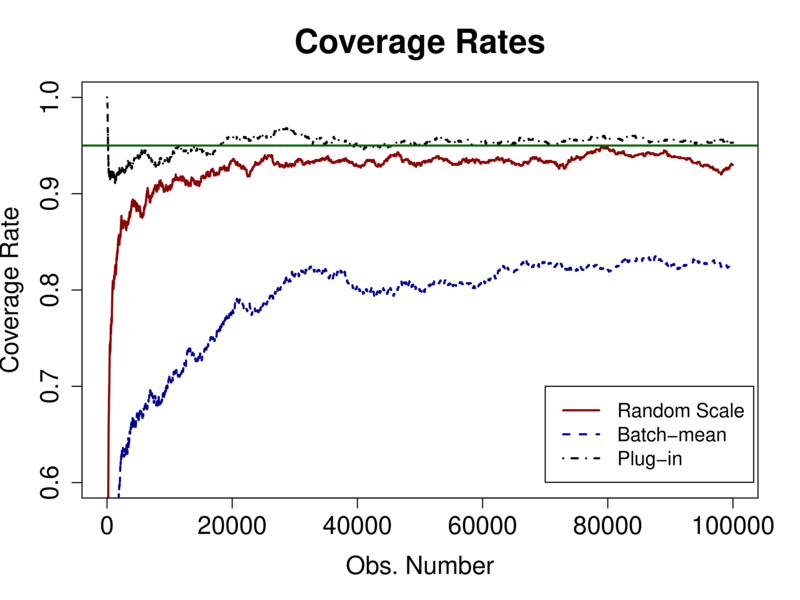}
\includegraphics[scale=0.15]{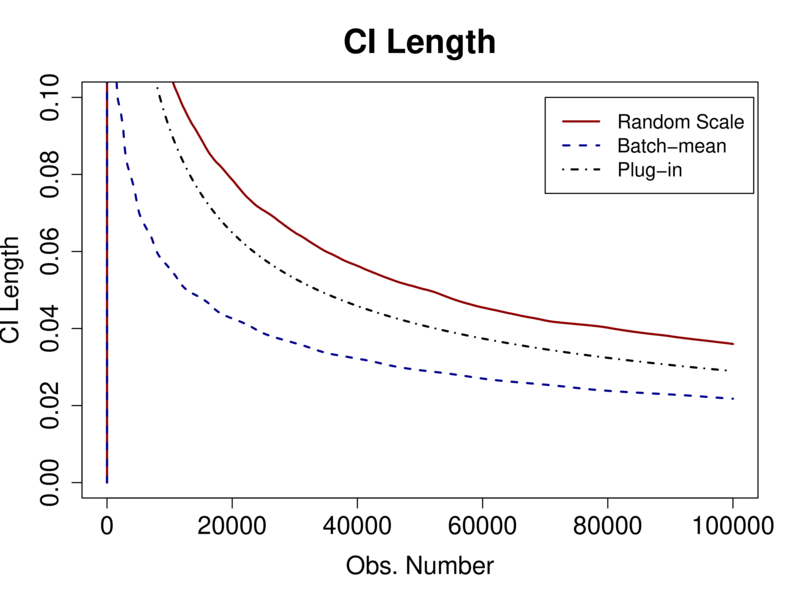}
\includegraphics[scale=0.15]{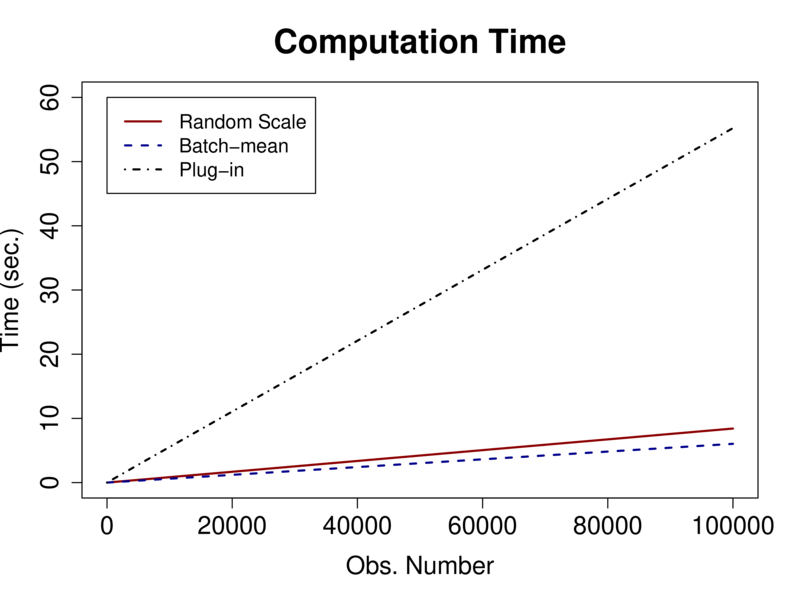}\\
\multicolumn{1}{c}{\underline{$\gamma_0=0.5, a=0.667$}}\\
\includegraphics[scale=0.15]{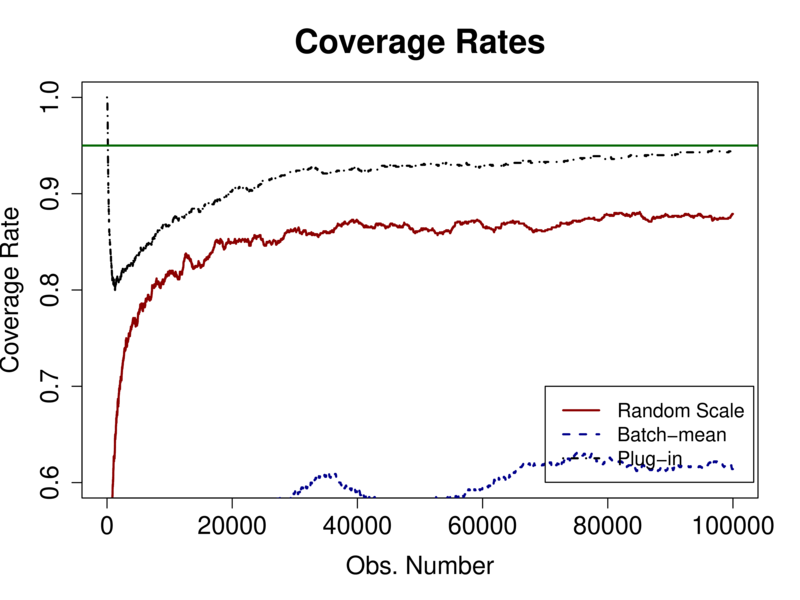}
\includegraphics[scale=0.15]{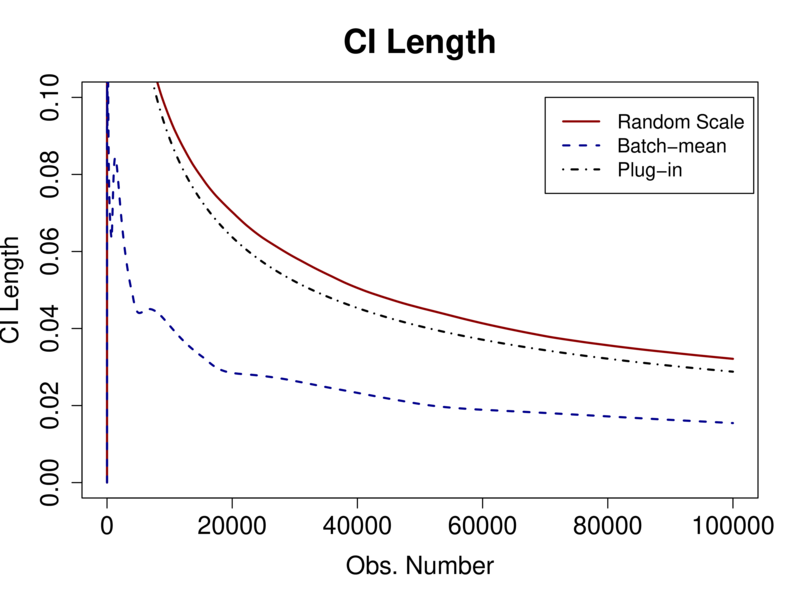}
\includegraphics[scale=0.15]{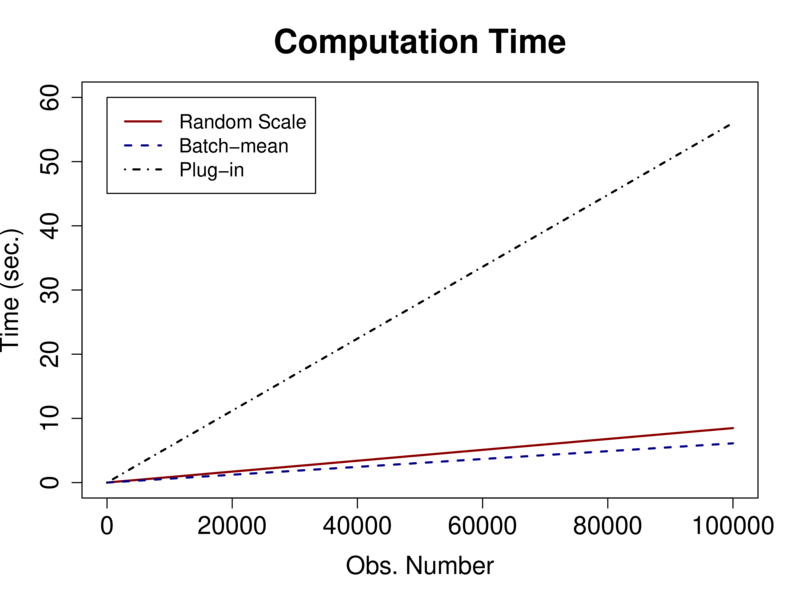}\\
\multicolumn{1}{c}{\underline{$\gamma_0=1, a=0.505$}}\\
\includegraphics[scale=0.15]{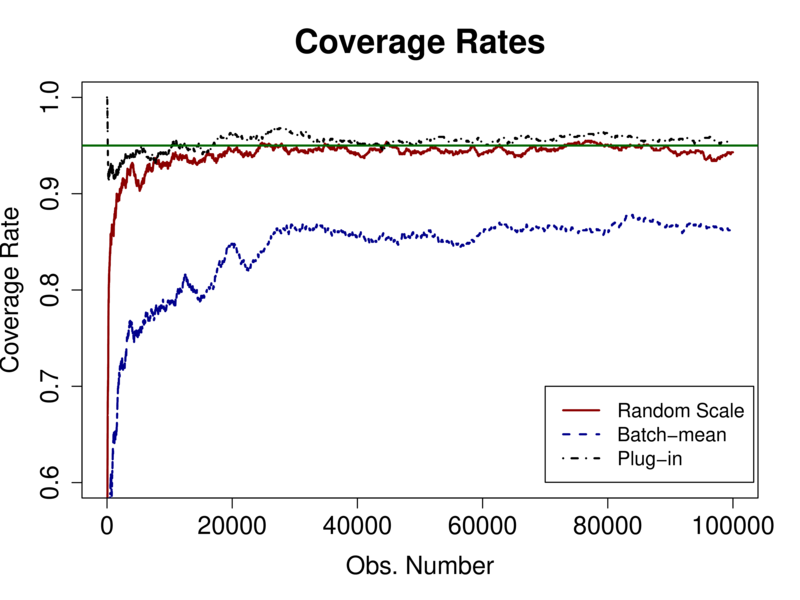}
\includegraphics[scale=0.15]{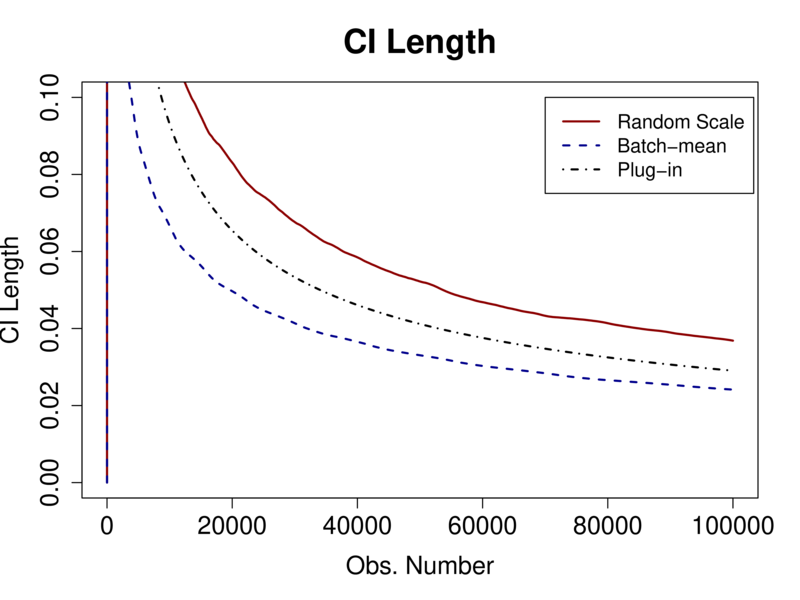}
\includegraphics[scale=0.15]{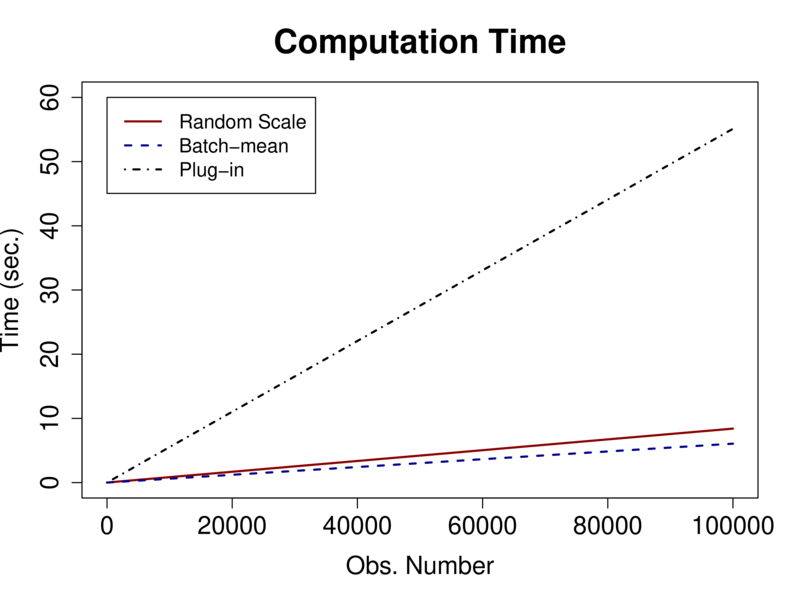}\\
\multicolumn{1}{c}{\underline{$\gamma_0=1, a=0.667$}}\\
\includegraphics[scale=0.15]{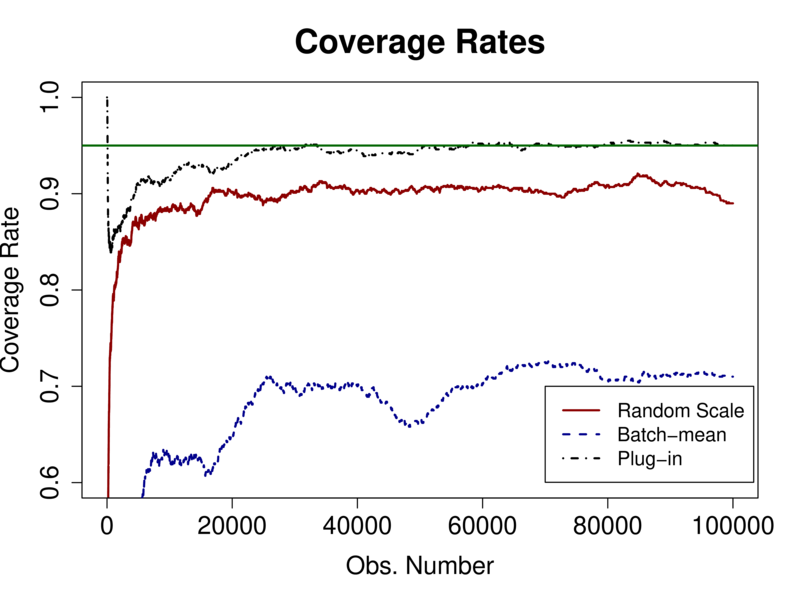}
\includegraphics[scale=0.15]{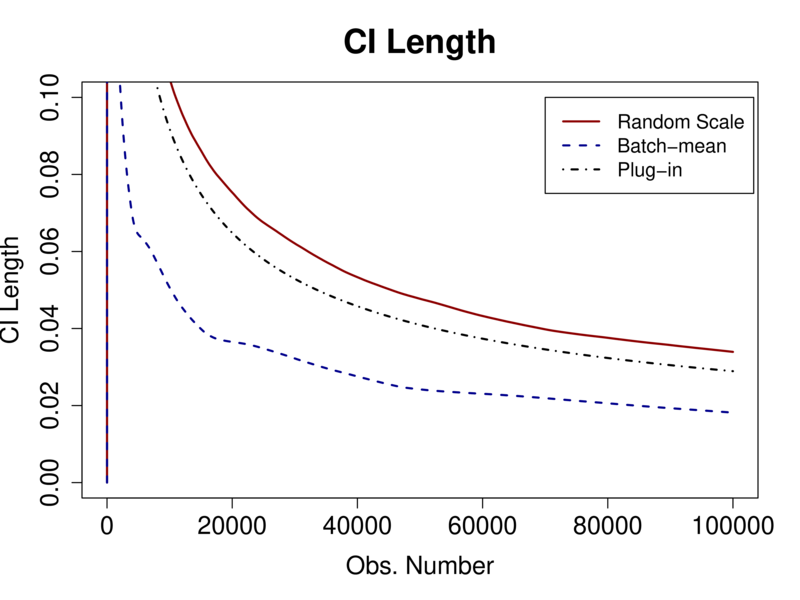}
\includegraphics[scale=0.15]{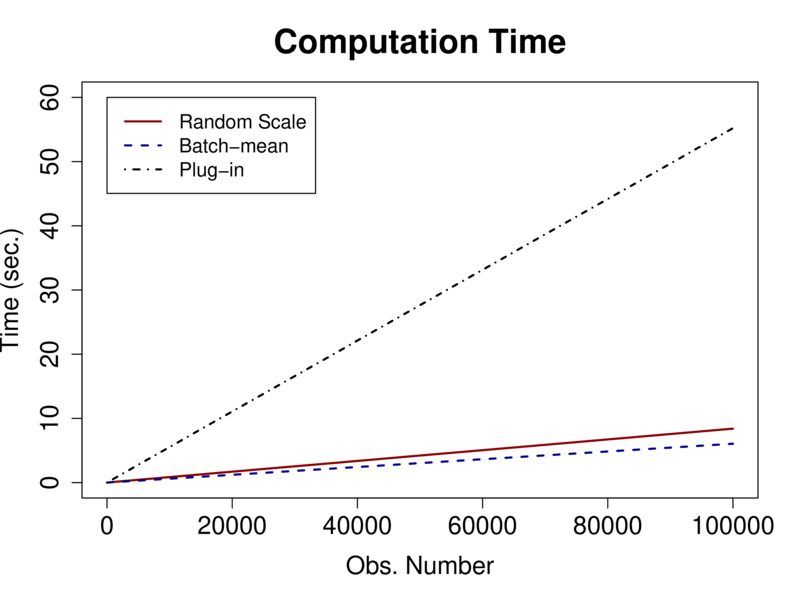}\\
\end{tabular}
\end{figure}

\begin{figure}[htp]
\caption{Logistic Model: $d=20$, $\gamma_0\in\{0.5, 1\}$, $a\in\{0.505, 0.667\}$  for $\gamma_t=\gamma_0 t^{-a}$}\label{fig_logit:d20}
\centering
\begin{tabular}{c}
\\
\multicolumn{1}{c}{\underline{$\gamma_0=0.5, a=0.505$}}\\
\includegraphics[scale=0.15]{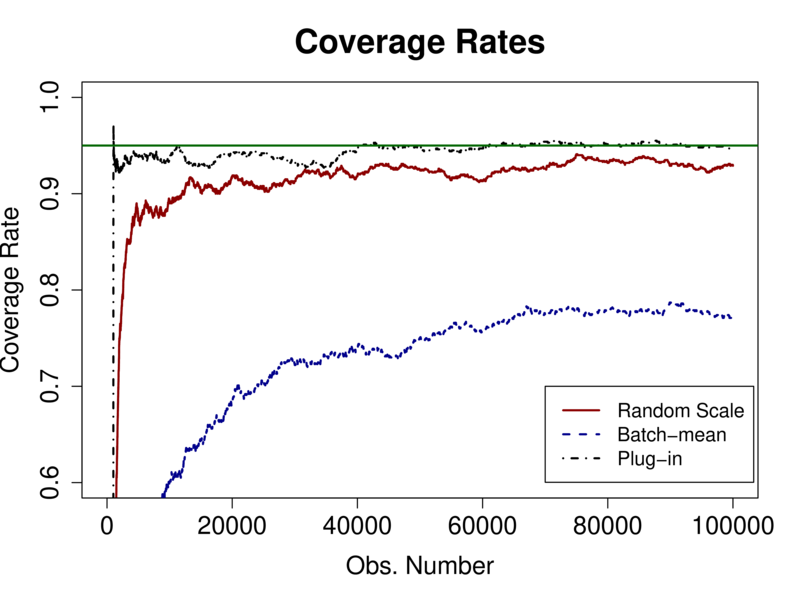}
\includegraphics[scale=0.15]{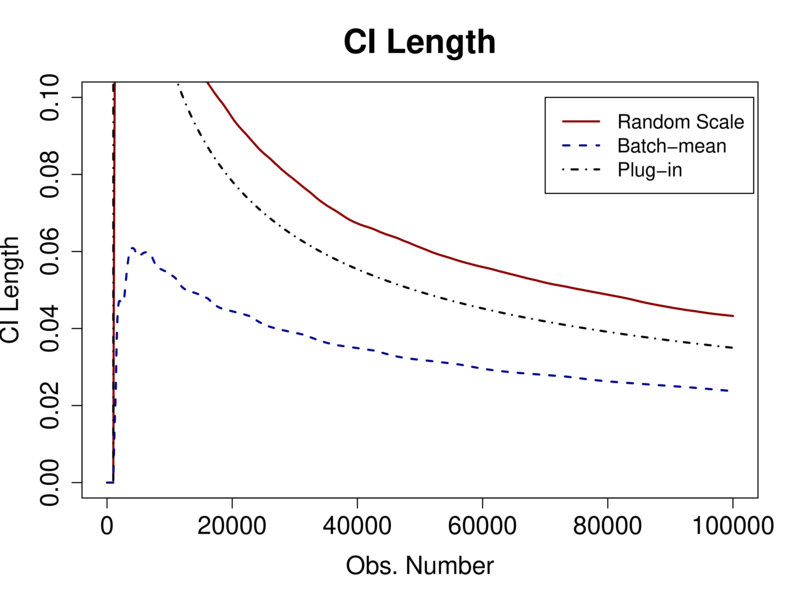}
\includegraphics[scale=0.15]{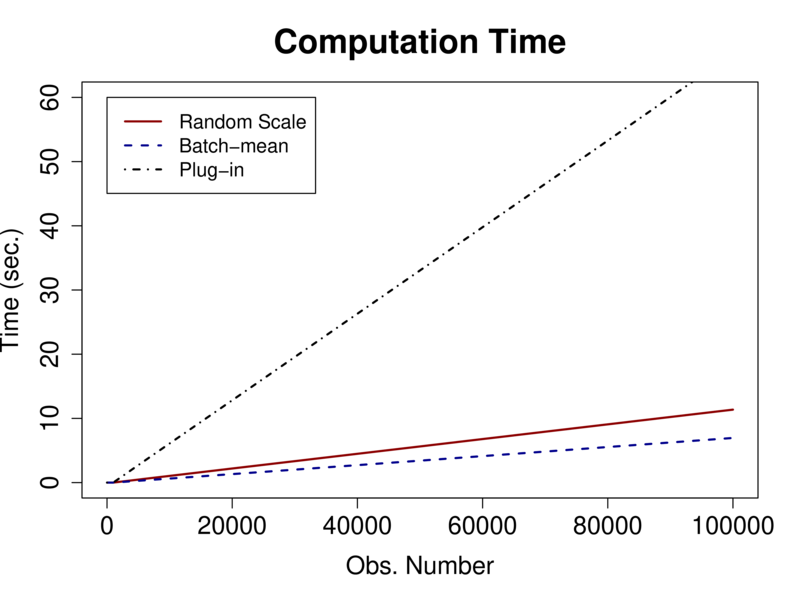}\\
\\
\multicolumn{1}{c}{\underline{$\gamma_0=0.5, a=0.667$}}\\
\includegraphics[scale=0.15]{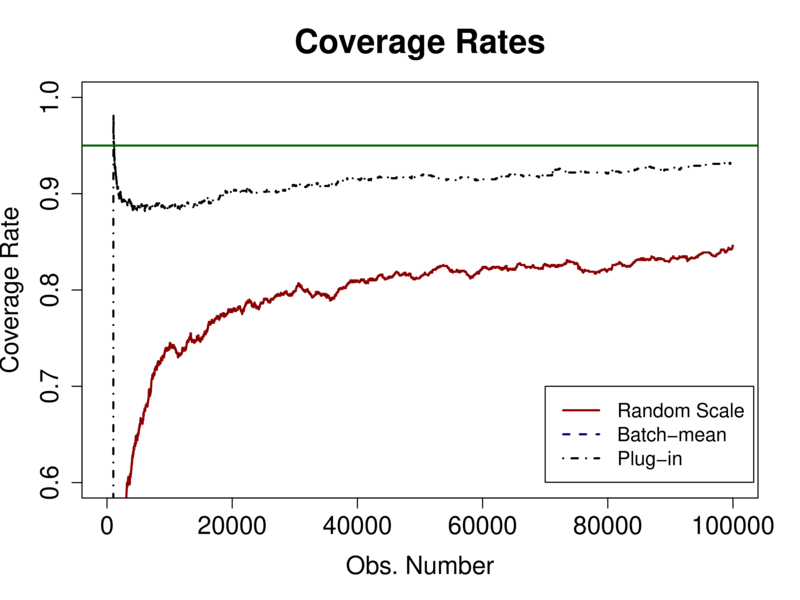}
\includegraphics[scale=0.15]{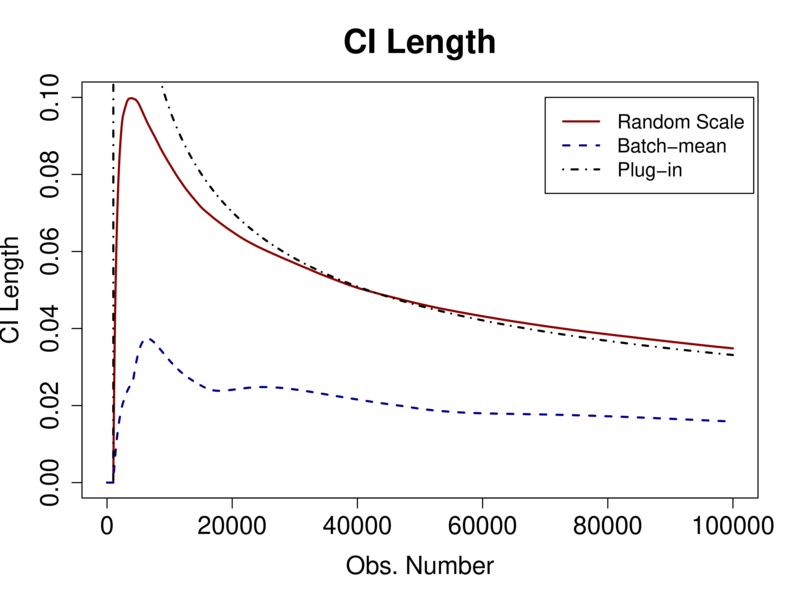}
\includegraphics[scale=0.15]{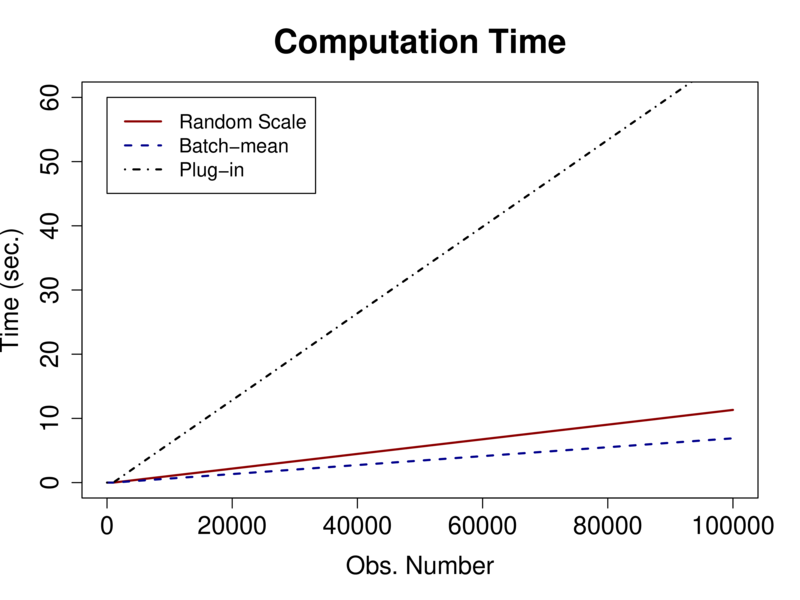}\\
\\
\multicolumn{1}{c}{\underline{$\gamma_0=1, a=0.505$}}\\
\includegraphics[scale=0.15]{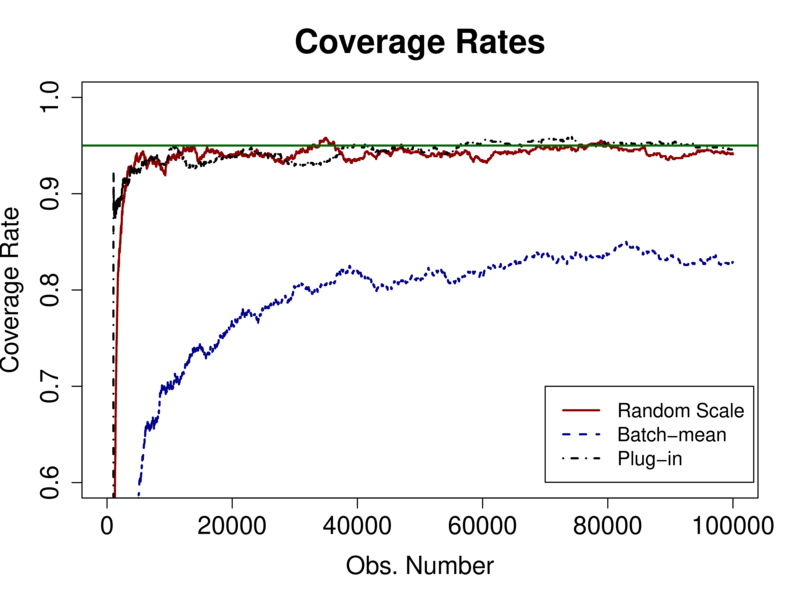}
\includegraphics[scale=0.15]{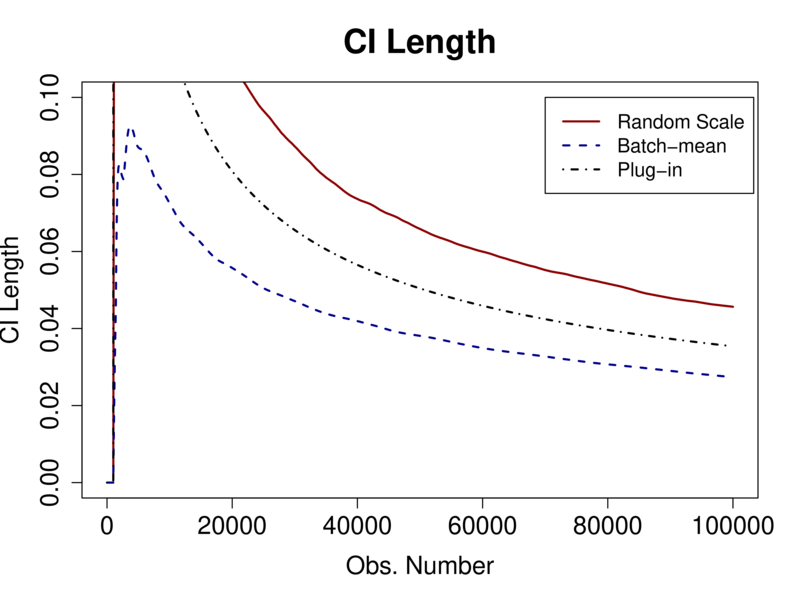}
\includegraphics[scale=0.15]{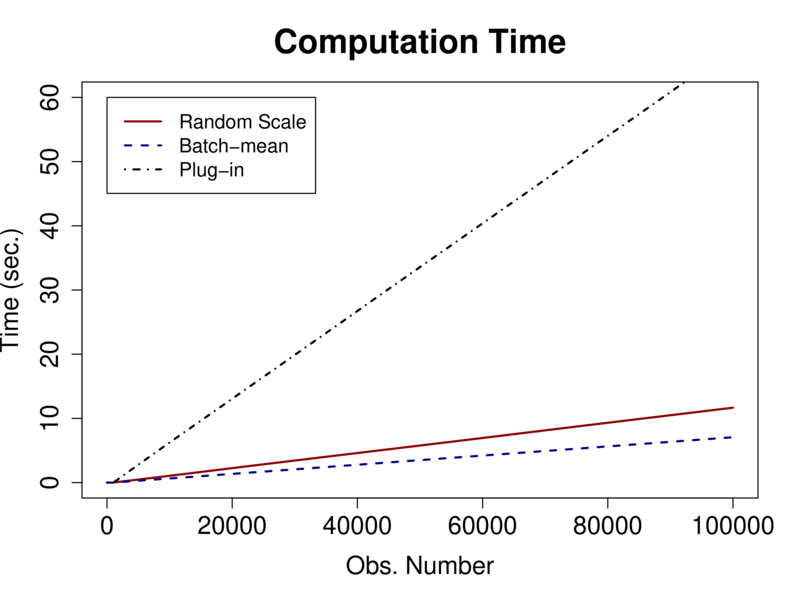}\\
\\
\multicolumn{1}{c}{\underline{$\gamma_0=1, a=0.667$}}\\
\includegraphics[scale=0.15]{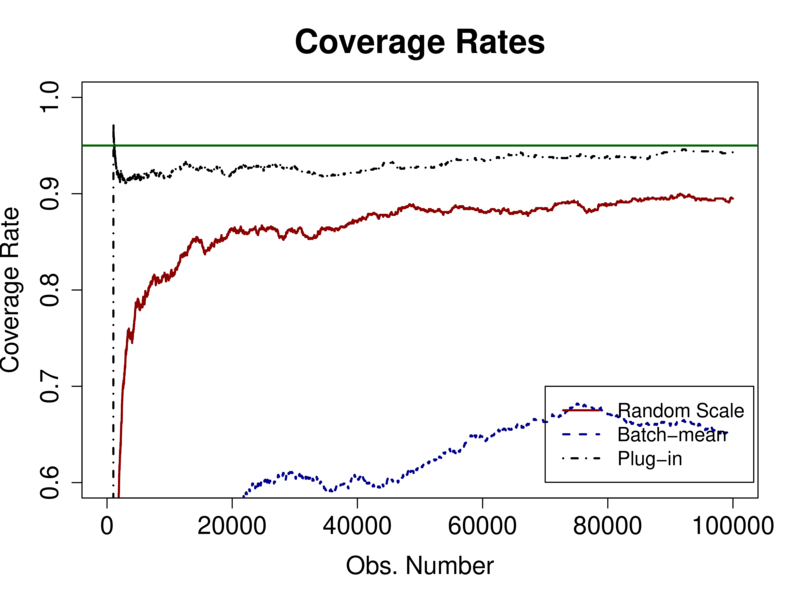}
\includegraphics[scale=0.15]{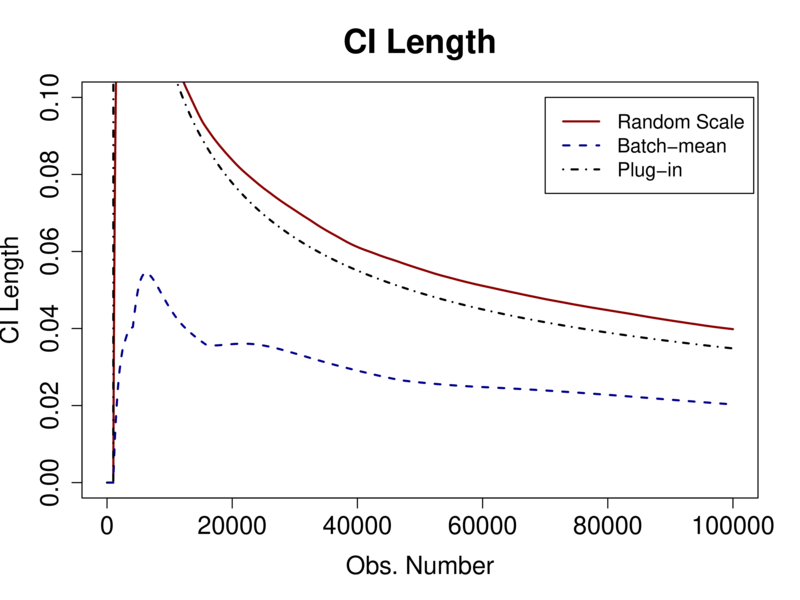}
\includegraphics[scale=0.15]{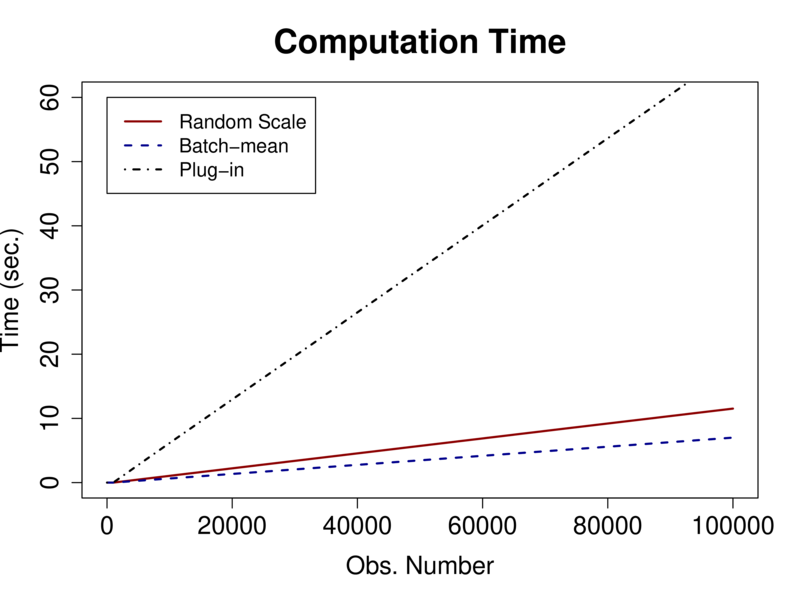}\\
\end{tabular}
\end{figure}

\begin{figure}[htp]
\caption{Logistic Model: $d=200$, $\gamma_0\in\{0.5, 1\}$, $a\in\{0.505, 0.667\}$  for $\gamma_t=\gamma_0 t^{-a}$}\label{fig_logit:d200}
\centering
\begin{tabular}{c}
\\
\multicolumn{1}{c}{\underline{$\gamma_0=0.5, a=0.505$}}\\
\includegraphics[scale=0.15]{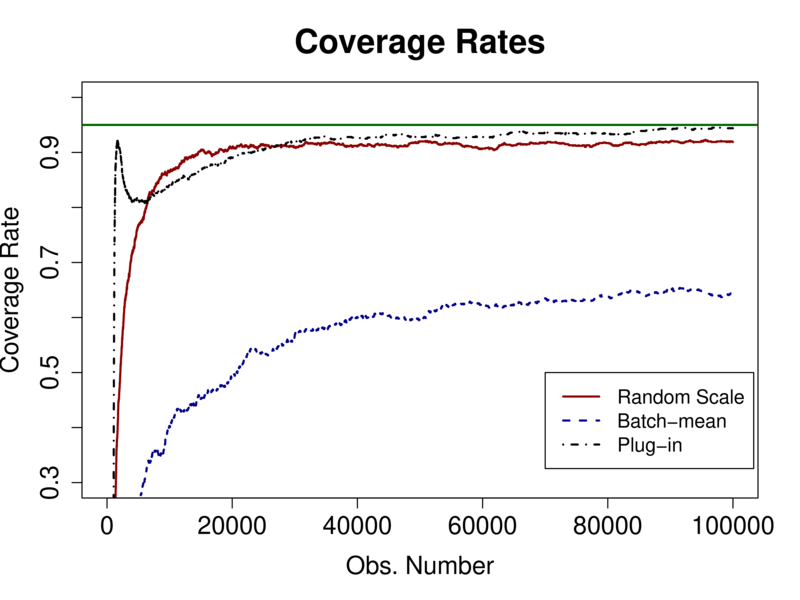}
\includegraphics[scale=0.15]{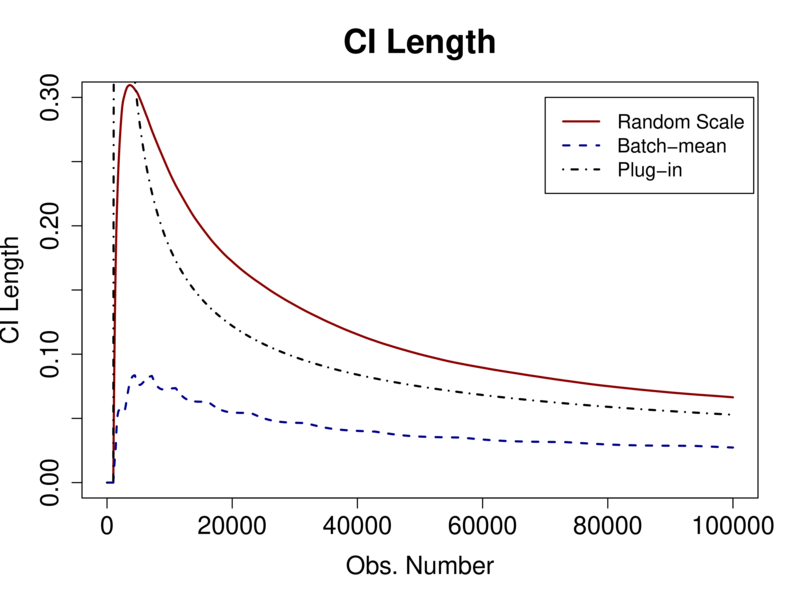}
\includegraphics[scale=0.15]{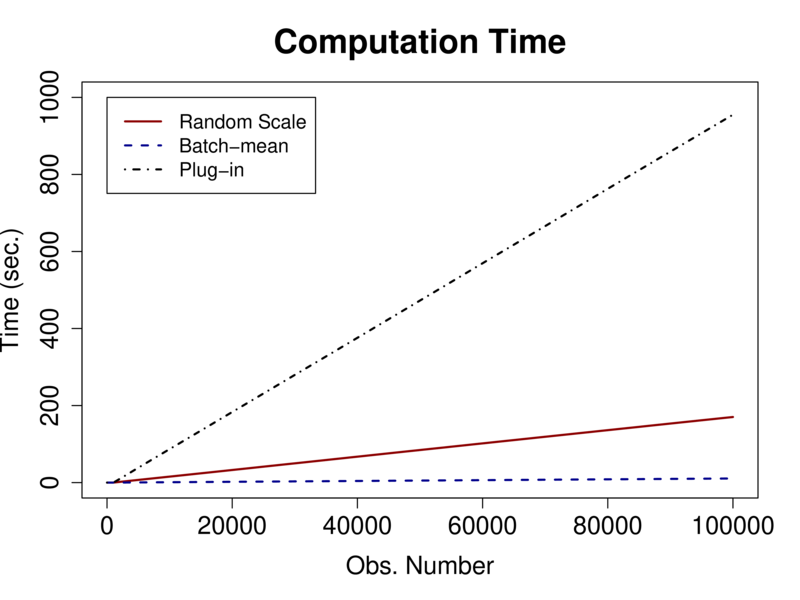}\\
\\
\multicolumn{1}{c}{\underline{$\gamma_0=0.5, a=0.667$}}\\
\includegraphics[scale=0.15]{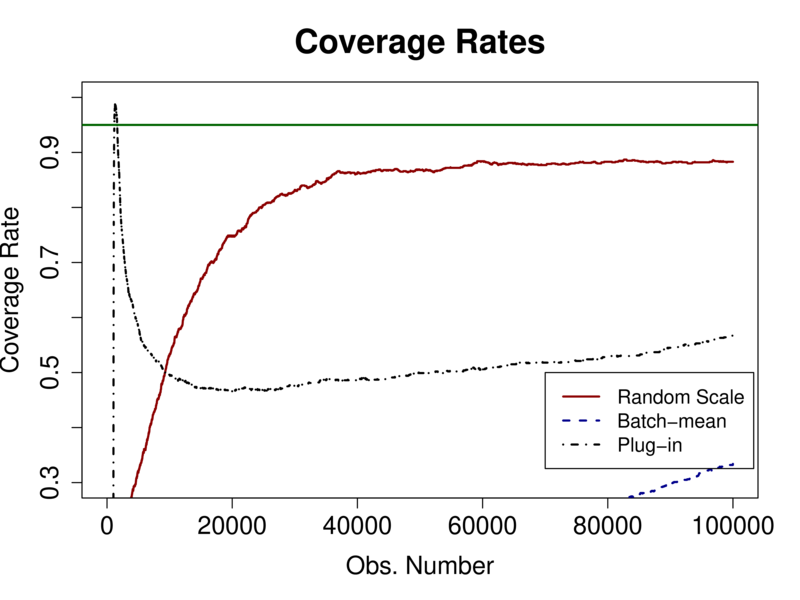}
\includegraphics[scale=0.15]{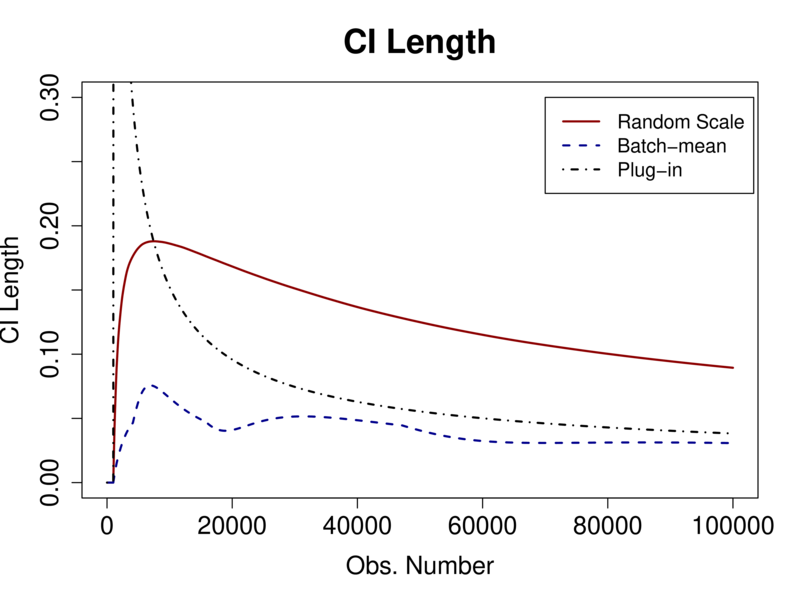}
\includegraphics[scale=0.15]{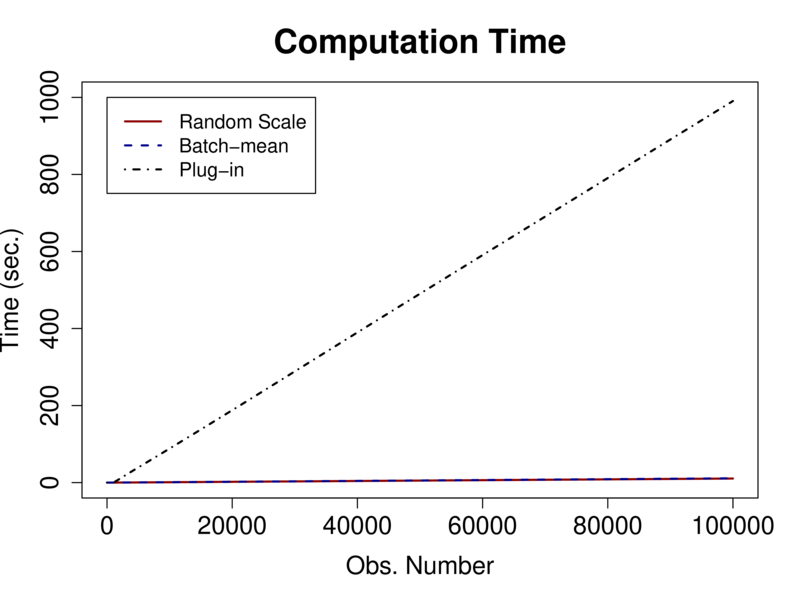}\\
\\
\multicolumn{1}{c}{\underline{$\gamma_0=1, a=0.505$}}\\
\includegraphics[scale=0.15]{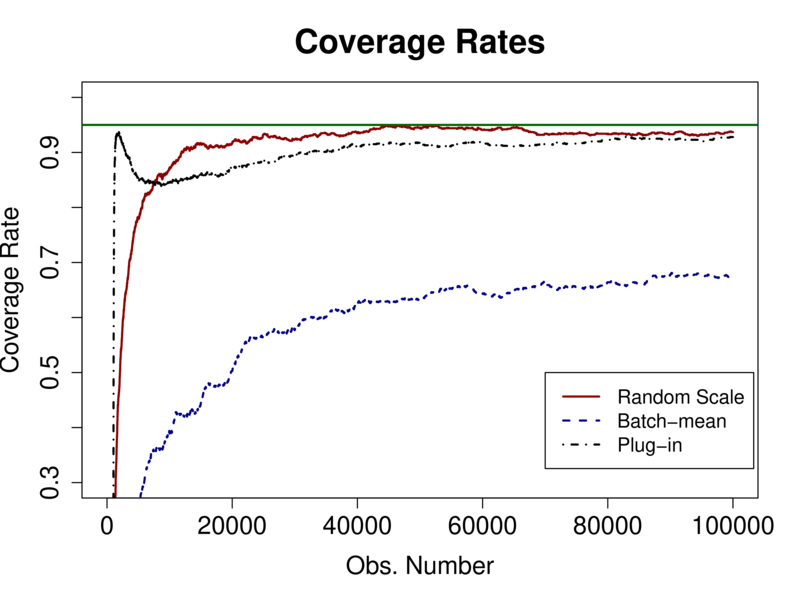}
\includegraphics[scale=0.15]{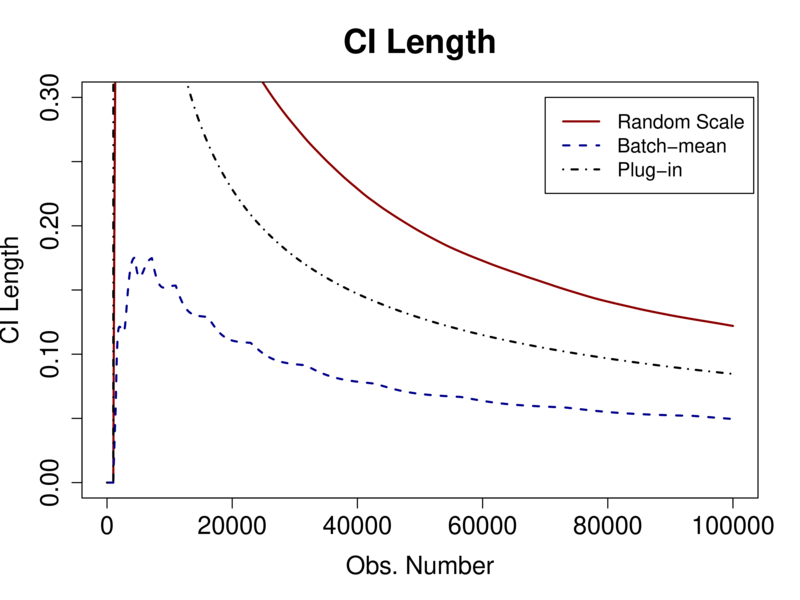}
\includegraphics[scale=0.15]{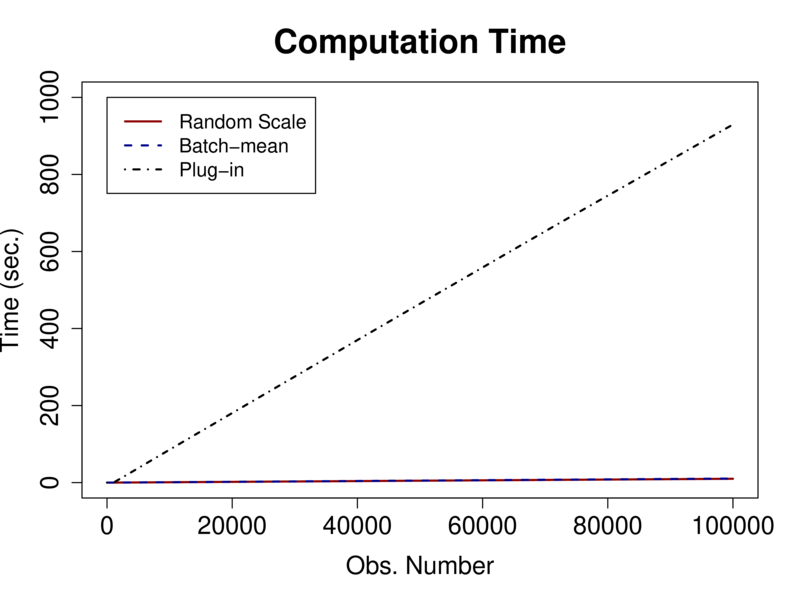}\\
\\
\multicolumn{1}{c}{\underline{$\gamma_0=1, a=0.667$}}\\
\includegraphics[scale=0.15]{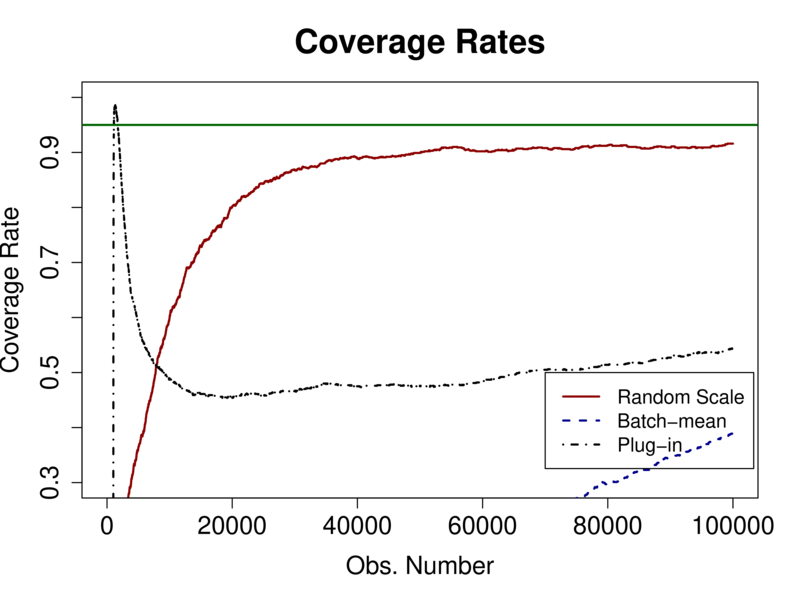}
\includegraphics[scale=0.15]{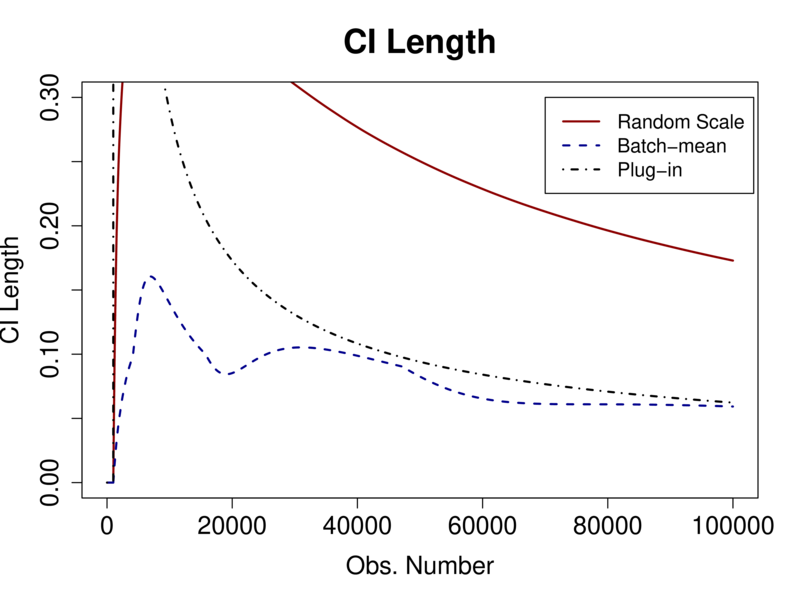}
\includegraphics[scale=0.15]{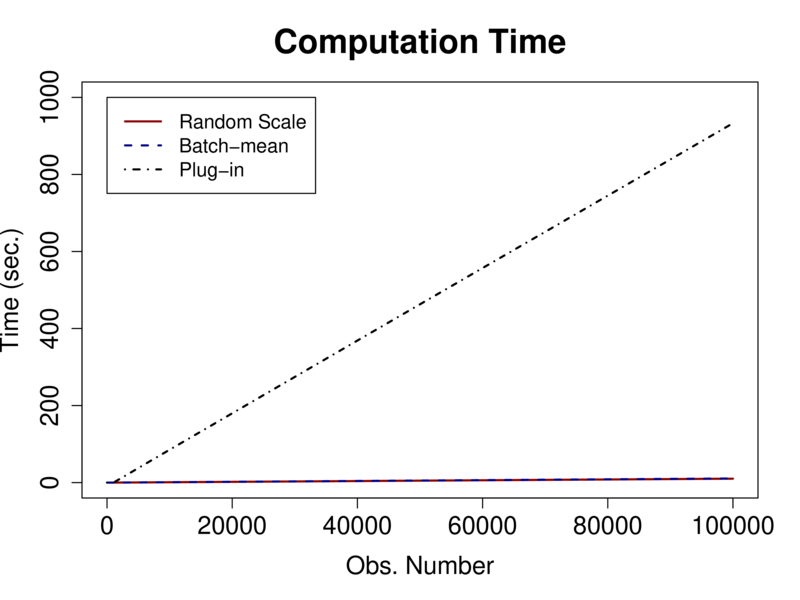}\\
\end{tabular}
\end{figure}

Figures~\ref{fig_logit:d5}--\ref{fig_logit:d200} show the complete paths of coverage rates, confidence interval lengths, and the computation times for each design with different learning rates. Tables~\ref{tb_logit:d=5}--\ref{tb_logit:d=200} report the same statistics at $n=25000, 50000, 75000,$ and $100000$. Overall, the results are similar to those of the linear regression model. The proposed random scaling method and the plug-in method fit the size well while the batch-mean method show less precise coverage rates. The computation time of the plug-in method is at least five times slower than the other two methods. Because the batch size depends on the learning rate parameter $a$, the batch-mean results are sensitive to different learning rates. For the large model $d=200$, the plug-in method does not perform well in terms of the coverage rate when $a=0.667$.

\begin{table}[ht]
\centering
\footnotesize
\caption{Logistic Model: $d=5$, $\gamma_0\in\{0.5, 1\}$, $a\in\{0.505, 0.667\}$  for $\gamma_t=\gamma_0 t^{-a}$}
\label{tb_logit:d=5}
\begin{tabular}{llcccc}
  \toprule
  & & $n=25000$ & $n=50000$ & $n=75000$ & $n=100000$ \\
  \midrule
  \multicolumn{3}{l}{\underline{$\gamma_0=0.5, a=0.505$}}\\
  \multirow{3}{*}{Coverage} 
  & Random Scale & 0.937 (0.0077) & 0.930 (0.0081) & 0.939 (0.0076) & 0.930 (0.0081) \\ 
  & Batch-mean   & 0.786 (0.0130) & 0.799 (0.0127) & 0.821 (0.0121) & 0.824 (0.0120) \\ 
  & Plug-in      & 0.960 (0.0062) & 0.951 (0.0068) & 0.955 (0.0066) & 0.953 (0.0067) \\ 
  \hline
  \multirow{3}{*}{Length} 
  & Random Scale & 0.071 & 0.050 & 0.041 & 0.036 \\ 
  & Batch-mean   & 0.039 & 0.029 & 0.025 & 0.022 \\ 
  & Plug-in      & 0.058 & 0.041 & 0.033 & 0.029 \\ 
  \hline
  \multirow{3}{*}{Time (sec.)} 
  & Random Scale & 2.1 & 4.2 & 6.3 & 8.4 \\ 
  & Batch-mean   & 1.5 & 3.0 & 4.5 & 6.0 \\ 
  & Plug-in      & 13.8 & 27.6 & 41.4 & 55.2 \\ 
  \hline
  \\
  \multicolumn{3}{l}{\underline{$\gamma_0=0.5, a=0.667$}}\\
  \multirow{3}{*}{Coverage} 
  & Random Scale & 0.848 (0.0114) & 0.862 (0.0109) & 0.871 (0.0106) & 0.879 (0.0103) \\ 
  & Batch-mean   & 0.566 (0.0157) & 0.578 (0.0156) & 0.630 (0.0153) & 0.614 (0.0154) \\ 
  & Plug-in      & 0.914 (0.0089) & 0.930 (0.0081) & 0.936 (0.0077) & 0.945 (0.0072) \\ 
  \hline
  \multirow{3}{*}{Length} 
  & Random Scale & 0.063 & 0.045 & 0.037 & 0.032 \\ 
  & Batch-mean   & 0.028 & 0.020 & 0.018 & 0.015 \\ 
  & Plug-in      & 0.057 & 0.041 & 0.033 & 0.029 \\ 
  \hline
  \multirow{3}{*}{Time (sec.)} 
  & Random Scale & 2.1 & 4.3 & 6.4 & 8.5 \\ 
  & Batch-mean   & 1.5 & 3.1 & 4.6 & 6.1 \\ 
  & Plug-in      & 14.0 & 28.0 & 42.0 & 56.0 \\ 
  \hline
  \\
  \multicolumn{3}{l}{\underline{$\gamma_0=1, a=0.505$}}\\
  \multirow{3}{*}{Coverage} 
  & Random Scale & 0.952 (0.0068) & 0.942 (0.0074) & 0.952 (0.0068) & 0.943 (0.0073) \\ 
  & Batch-mean   & 0.841 (0.0116) & 0.855 (0.0111) & 0.862 (0.0109) & 0.863 (0.0109) \\ 
  & Plug-in      & 0.959 (0.0063) & 0.953 (0.0067) & 0.959 (0.0063) & 0.952 (0.0068) \\ 
  \hline 
  \multirow{3}{*}{Length} 
  & Random Scale & 0.074 & 0.052 & 0.042 & 0.037 \\ 
  & Batch-mean   & 0.045 & 0.033 & 0.027 & 0.024 \\ 
  & Plug-in      & 0.058 & 0.041 & 0.034 & 0.029 \\ 
  \hline
  \multirow{3}{*}{Time (sec.)} 
  & Random Scale & 2.1 & 4.2 & 6.3 & 8.4 \\ 
  & Batch-mean   & 1.5 & 3.0 & 4.5 & 6.1 \\ 
  & Plug-in      & 13.8 & 27.6 & 41.3 & 55.1 \\
  \hline
   \\
  \multicolumn{3}{l}{\underline{$\gamma_0=1, a=0.667$}}\\
  \multirow{3}{*}{Coverage} 
  & Random Scale & 0.892 (0.0098) & 0.904 (0.0093) & 0.904 (0.0093) & 0.890 (0.0099) \\ 
  & Batch-mean   & 0.702 (0.0145) & 0.662 (0.0150) & 0.722 (0.0142) & 0.710 (0.0143) \\ 
  & Plug-in      & 0.945 (0.0072) & 0.946 (0.0071) & 0.949 (0.0070) & 0.950 (0.0069) \\ 
  \hline 
  \multirow{3}{*}{Length} 
  & Random Scale & 0.068 & 0.048 & 0.039 & 0.034 \\ 
  & Batch-mean   & 0.035 & 0.024 & 0.021 & 0.018 \\ 
  & Plug-in      & 0.058 & 0.041 & 0.033 & 0.029 \\ 
  \hline
  \multirow{3}{*}{Time (sec.)} 
  & Random Scale & 2.1 & 4.2 & 6.3 & 8.4 \\ 
  & Batch-mean   & 1.5 & 3.0 & 4.5 & 6.1 \\ 
  & Plug-in      & 13.8 & 27.6 & 41.4 & 55.2 \\ 
  \bottomrule
\end{tabular}
\end{table}

\begin{table}[ht]
\centering
\footnotesize
\caption{Logistic Model: $d=20$, $\gamma_0\in\{0.5, 1\}$, $a\in\{0.505, 0.667\}$  for $\gamma_t=\gamma_0 t^{-a}$}
\label{tb_logit:d=20}
\begin{tabular}{llcccc}
  \toprule
  & & $n=25000$ & $n=50000$ & $n=75000$ & $n=100000$ \\
 \midrule
  \multicolumn{3}{l}{\underline{$\gamma_0=0.5, a=0.505$}}\\
  \multirow{3}{*}{Coverage} 
  & Random Scale & 0.908 (0.0091) & 0.925 (0.0083) & 0.941 (0.0075) & 0.929 (0.0081) \\ 
  & Batch-mean   & 0.704 (0.0144) & 0.750 (0.0137) & 0.780 (0.0131) & 0.772 (0.0133) \\ 
  & Plug-in      & 0.942 (0.0074) & 0.947 (0.0071) & 0.953 (0.0067) & 0.946 (0.0071) \\ 
  \hline
  \multirow{3}{*}{Length} 
  & Random Scale & 0.086 & 0.061 & 0.050 & 0.043 \\ 
  & Batch-mean   & 0.041 & 0.032 & 0.027 & 0.024 \\ 
  & Plug-in      & 0.070 & 0.050 & 0.040 & 0.035 \\ 
  \hline
  \multirow{3}{*}{Time (sec.)} 
  & Random Scale & 2.8 & 5.6 & 8.5 & 11.4 \\ 
  & Batch-mean   & 1.7 & 3.4 & 5.2 & 7.0 \\ 
  & Plug-in      & 16.2 & 33.0 & 49.9 & 66.8 \\
  \hline
  \\
  \multicolumn{3}{l}{\underline{$\gamma_0=0.5, a=0.667$}}\\
  \multirow{3}{*}{Coverage} 
  & Random Scale & 0.788 (0.0129) & 0.813 (0.0123) & 0.828 (0.0119) & 0.846 (0.0114) \\ 
  & Batch-mean   & 0.456 (0.0158) & 0.480 (0.0158) & 0.527 (0.0158) & 0.549 (0.0157) \\ 
  & Plug-in      & 0.902 (0.0094) & 0.919 (0.0086) & 0.922 (0.0085) & 0.932 (0.0080) \\ 
  \hline
  \multirow{3}{*}{Length} 
  & Random Scale & 0.061 & 0.046 & 0.039 & 0.035 \\ 
  & Batch-mean   & 0.025 & 0.019 & 0.017 & 0.016 \\ 
  & Plug-in      & 0.063 & 0.046 & 0.038 & 0.033 \\ 
  \hline
  \multirow{3}{*}{Time (sec.)} 
  & Random Scale & 2.8 & 5.6 & 8.5 & 11.3 \\ 
  & Batch-mean   & 1.7 & 3.4 & 5.2 & 6.9 \\ 
  & Plug-in      & 16.2 & 33.1 & 50.0 & 66.9 \\ 
  \hline
  \\
  \multicolumn{3}{l}{\underline{$\gamma_0=1, a=0.505$}}\\
  \multirow{3}{*}{Coverage} 
  & Random Scale & 0.939 (0.0076) & 0.942 (0.0074) & 0.945 (0.0072) & 0.941 (0.0075) \\ 
  & Batch-mean   & 0.782 (0.0131) & 0.812 (0.0124) & 0.838 (0.0117) & 0.829 (0.0119) \\ 
  & Plug-in      & 0.942 (0.0074) & 0.947 (0.0071) & 0.952 (0.0068) & 0.943 (0.0073) \\ 
  \hline 
  \multirow{3}{*}{Length} 
  & Random Scale & 0.097 & 0.066 & 0.053 & 0.046 \\ 
  & Batch-mean   & 0.050 & 0.038 & 0.032 & 0.027 \\ 
  & Plug-in      & 0.072 & 0.050 & 0.041 & 0.035 \\ 
  \hline
  \multirow{3}{*}{Time (sec.)} 
  & Random Scale & 2.8 & 5.8 & 8.7 & 11.7 \\ 
  & Batch-mean   & 1.7 & 3.5 & 5.3 & 7.1 \\ 
  & Plug-in      & 16.5 & 33.6 & 50.6 & 67.7 \\ 
  \hline
   \\
  \multicolumn{3}{l}{\underline{$\gamma_0=1, a=0.667$}}\\
  \multirow{3}{*}{Coverage} 
  & Random Scale & 0.863 (0.0109) & 0.885 (0.0101) & 0.891 (0.0099) & 0.895 (0.0097) \\ 
  & Batch-mean   & 0.604 (0.0155) & 0.614 (0.0154) & 0.681 (0.0147) & 0.653 (0.0151) \\ 
  & Plug-in      & 0.924 (0.0084) & 0.928 (0.0082) & 0.939 (0.0076) & 0.944 (0.0073) \\ 
  \hline 
  \multirow{3}{*}{Length} 
  & Random Scale & 0.076 & 0.055 & 0.046 & 0.040 \\ 
  & Batch-mean   & 0.036 & 0.026 & 0.023 & 0.020 \\ 
  & Plug-in      & 0.070 & 0.049 & 0.040 & 0.035 \\ 
  \hline
  \multirow{3}{*}{Time (sec.)} 
  & Random Scale & 2.8 & 5.7 & 8.6 & 11.5 \\ 
  & Batch-mean   & 1.7 & 3.5 & 5.2 & 7.0 \\ 
  & Plug-in      & 16.3 & 33.3 & 50.2 & 67.2 \\ 
  \bottomrule
\end{tabular}
\end{table}

\begin{table}[ht]
\centering
\footnotesize
\caption{Logistic Model: $d=200$, $\gamma_0\in\{0.5, 1\}$, $a\in\{0.505, 0.667\}$  for $\gamma_t=\gamma_0 t^{-a}$}
\label{tb_logit:d=200}
\begin{tabular}{llcccc}
  \toprule
  & & $n=25000$ & $n=50000$ & $n=75000$ & $n=100000$ \\
  \midrule
  \multicolumn{3}{l}{\underline{$\gamma_0=0.5, a=0.505$}}\\
  \multirow{3}{*}{Coverage} 
  & Random Scale & 0.914 (0.0089) & 0.920 (0.0086) & 0.912 (0.0090) & 0.919 (0.0086) \\  
  & Batch-mean   & 0.534 (0.0158) & 0.594 (0.0155) & 0.630 (0.0153) & 0.644 (0.0151) \\ 
  & Plug-in      & 0.904 (0.0093) & 0.928 (0.0082) & 0.935 (0.0078) & 0.944 (0.0073) \\ 
  \hline
  \multirow{3}{*}{Length} 
  & Random Scale & 0.153 & 0.100 & 0.078 & 0.066 \\ 
  & Batch-mean   & 0.050 & 0.036 & 0.031 & 0.027 \\ 
  & Plug-in      & 0.108 & 0.075 & 0.061 & 0.053 \\ 
  \hline
  \multirow{3}{*}{Time (sec.)} 
  & Random Scale & 41.4 & 84.6 & 127.6 & 170.3 \\ 
  & Batch-mean   & 2.6 & 5.3 & 8.0 & 10.7 \\ 
  & Plug-in      & 231.4 & 472.6 & 714.3 & 955.0 \\ 
  \hline
  \\
  \multicolumn{3}{l}{\underline{$\gamma_0=0.5, a=0.667$}}\\
  \multirow{3}{*}{Coverage} 
  & Random Scale & 0.804 (0.0126) & 0.868 (0.0107) & 0.881 (0.0102) & 0.883 (0.0102) \\ 
  & Batch-mean   & 0.165 (0.0117) & 0.216 (0.0130) & 0.244 (0.0136) & 0.334 (0.0149) \\ 
  & Plug-in      & 0.467 (0.0158) & 0.499 (0.0158) & 0.521 (0.0158) & 0.567 (0.0157) \\ 
  \hline
  \multirow{3}{*}{Length} 
  & Random Scale & 0.159 & 0.125 & 0.104 & 0.089 \\ 
  & Batch-mean   & 0.048 & 0.041 & 0.031 & 0.031 \\ 
  & Plug-in      & 0.083 & 0.055 & 0.044 & 0.038 \\ 
  \hline
  \multirow{3}{*}{Time (sec.)} 
  & Random Scale & 2.6 & 5.2 & 7.9 & 10.5 \\ 
  & Batch-mean   & 2.7 & 5.5 & 8.3 & 11.1 \\ 
  & Plug-in      & 238.2 & 490.3 & 740.6 & 990.6 \\ 
  \hline
  \\
  \multicolumn{3}{l}{\underline{$\gamma_0=1, a=0.505$}}\\
  \multirow{3}{*}{Coverage} 
  & Random Scale & 0.933 (0.0079) & 0.944 (0.0073) & 0.934 (0.0079) & 0.937 (0.0077) \\ 
  & Batch-mean   & 0.566 (0.0157) & 0.631 (0.0153) & 0.656 (0.0150) & 0.674 (0.0148) \\ 
  & Plug-in      & 0.884 (0.0101) & 0.916 (0.0088) & 0.917 (0.0087) & 0.928 (0.0082) \\ 
  \hline 
  \multirow{3}{*}{Length} 
  & Random Scale & 0.311 & 0.196 & 0.148 & 0.122 \\ 
  & Batch-mean   & 0.100 & 0.069 & 0.058 & 0.050 \\ 
  & Plug-in      & 0.197 & 0.128 & 0.101 & 0.085 \\ 
  \hline
  \multirow{3}{*}{Time (sec.)} 
  & Random Scale & 2.4 & 4.9 & 7.4 & 9.9 \\  
  & Batch-mean   & 2.5 & 5.2 & 7.8 & 10.5 \\ 
  & Plug-in      & 227.9 & 464.5 & 698.8 & 929.6 \\ 
  \hline
   \\
  \multicolumn{3}{l}{\underline{$\gamma_0=1, a=0.667$}}\\
  \multirow{3}{*}{Coverage} 
  & Random Scale & 0.844 (0.0115) & 0.899 (0.0095) & 0.908 (0.0091) & 0.916 (0.0088) \\ 
  & Batch-mean   & 0.166 (0.0118) & 0.257 (0.0138) & 0.272 (0.0141) & 0.389 (0.0154) \\ 
  & Plug-in      & 0.458 (0.0158) & 0.475 (0.0158) & 0.505 (0.0158) & 0.543 (0.0158) \\ 
  \hline 
  \multirow{3}{*}{Length} 
  & Random Scale & 0.329 & 0.250 & 0.203 & 0.173 \\ 
  & Batch-mean   & 0.099 & 0.083 & 0.061 & 0.059 \\ 
  & Plug-in      & 0.148 & 0.094 & 0.074 & 0.062 \\ 
  \hline
  \multirow{3}{*}{Time (sec.)} 
  & Random Scale & 2.5 & 5.0 & 7.6 & 10.1 \\ 
  & Batch-mean   & 2.6 & 5.3 & 8.0 & 10.7 \\ 
  & Plug-in      & 227.1 & 462.9 & 698.0 & 933.0 \\ 
  \bottomrule
\end{tabular}
\end{table}

\clearpage

\bibliographystyle{chicago}
\bibliography{LLSS_bib}

\end{document}